\documentclass{article} 
\usepackage{iclr2027_conference, times}
\iclrfinalcopy

\makeatletter
\def\@notice{}

\makeatother

\usepackage{etoolbox}
\AtBeginDocument{%
  \fancypagestyle{plain}{%
    \fancyhf{}
    \renewcommand{\headrulewidth}{0pt}
  }%
}

\usepackage{amsmath,amsfonts,bm}

\def\eqref#1{equation~\ref{#1}}

\def\1{\bm{1}}

\DeclareMathAlphabet{\mathsfit}{\encodingdefault}{\sfdefault}{m}{sl}
\SetMathAlphabet{\mathsfit}{bold}{\encodingdefault}{\sfdefault}{bx}{n}

\usepackage[hyphens]{url}
\usepackage{graphicx}
\usepackage{placeins}
\usepackage{caption}
\usepackage{newfloat}

\usepackage{amsmath}
\usepackage{amssymb}

\usepackage{booktabs}
\usepackage{tabularx}
\usepackage{array}
\usepackage{multirow}
\usepackage{makecell}

\usepackage{algorithm}
\usepackage{algorithmic}

\usepackage{listings}
\DeclareCaptionStyle{ruled}{labelfont=normalfont,labelsep=colon,strut=off}
\floatstyle{ruled}
\newfloat{listing}{tb}{lst}{}
\floatname{listing}{Listing}

\usepackage{hyperref}
\hypersetup{
    colorlinks=true,
    linkcolor=red,
    citecolor=blue,
    urlcolor=blue
}

\title{PersMem: Internalizing Personality into Dual-Pathway Memory for LLM Agents}

\author{
  Hanzhong Zhang\textsuperscript{1},
  Ziwei Xiang\textsuperscript{2},
  Weicheng Xie\textsuperscript{3},
  Shizhe Liu\textsuperscript{4},
  Siyang Song\textsuperscript{1}\thanks{Corresponding author.} \\
  \textsuperscript{1}Department of Computer Science, University of Exeter, United Kingdom \\
  \textsuperscript{2}School of Marxism, Nanjing University, China \\
  \textsuperscript{3}Shenzhen University, China \\
  \textsuperscript{4}Department of Computer Science, University of Oxford, United Kingdom \\
  \texttt{armihiabelliard@foxmail.com}, \texttt{rhodus\_boar@smail.nju.edu.cn}, \\
  \texttt{wcxie@szu.edu.cn}, \texttt{shizhe.liu@oriel.ox.ac.uk}, \texttt{S.Song@exeter.ac.uk}
}

\iclrfinalcopy 

\begin{document}

\maketitle
\begin{abstract}

The profile of an existing role-playing agent usually depends on the user pre-defined personality in a system prompt, whereas its memory processing pipeline, including the prioritisation of stored memories and the guidance of subsequent retrieval, remains largely independent of this personality. This separation not only frequently causes the agent's memory processing to be inconsistent with the pre-defined personality, but also makes it difficult to validate whether the behaviours expressed by the agent follow this personality. In this paper, we propose a novel memory processing strategy called Personality-Integrated Memory (PersMem), which integrates the pre-defined personality into the role-playing agent's memory processing pipeline, thereby making the agent's memory processing consistently personality-dependent. Specifically, our PersMem processes the agent's memory using novel four personality-dependent steps, where the pre-defined personality is mapped to operation-specific parameters controlling: (i) the affective appraisal annotating the emotion states of the given user input for the incoming reply; (ii) retention of previously stored memories along with the current input; (iii) passive affect-driven memory retrieval exploring memories that are similar with the user input in terms of their semantics and personality-guided emotions; and (iv) active goal-driven memory retrieval that refines and selects passively retrieved memories for the agent's reply. This ensures the agent's memory processing to be fully consistent with the pre-defined personality. Consequently, the consistency of the behaviours expressed by the agent with the pre-defined personality can be examined by inspecting the agent's memory-processing traces during human-agent interactions. We evaluate these personality-dependent differences in attachment and Big Five settings. PersMem exceeds the chance baseline for four-way attachment-personality classification by 23.1 percentage points. In Big Five dialogue comparisons, PersMem also achieves 67.5\% accuracy, 6.7 percentage points above a baseline using uniformly sampled memories. On CoSER, PersMem achieves an average score of 66.13, with scores of 69.33 for Character Fidelity and 84.33 for Storyline Quality. Together, these results show that PersMem produces distinguishable personality-related memory-processing patterns.

\end{abstract}

\section{Introduction}

\begin{figure*}[t]
  \centering
  \includegraphics[width=\textwidth]{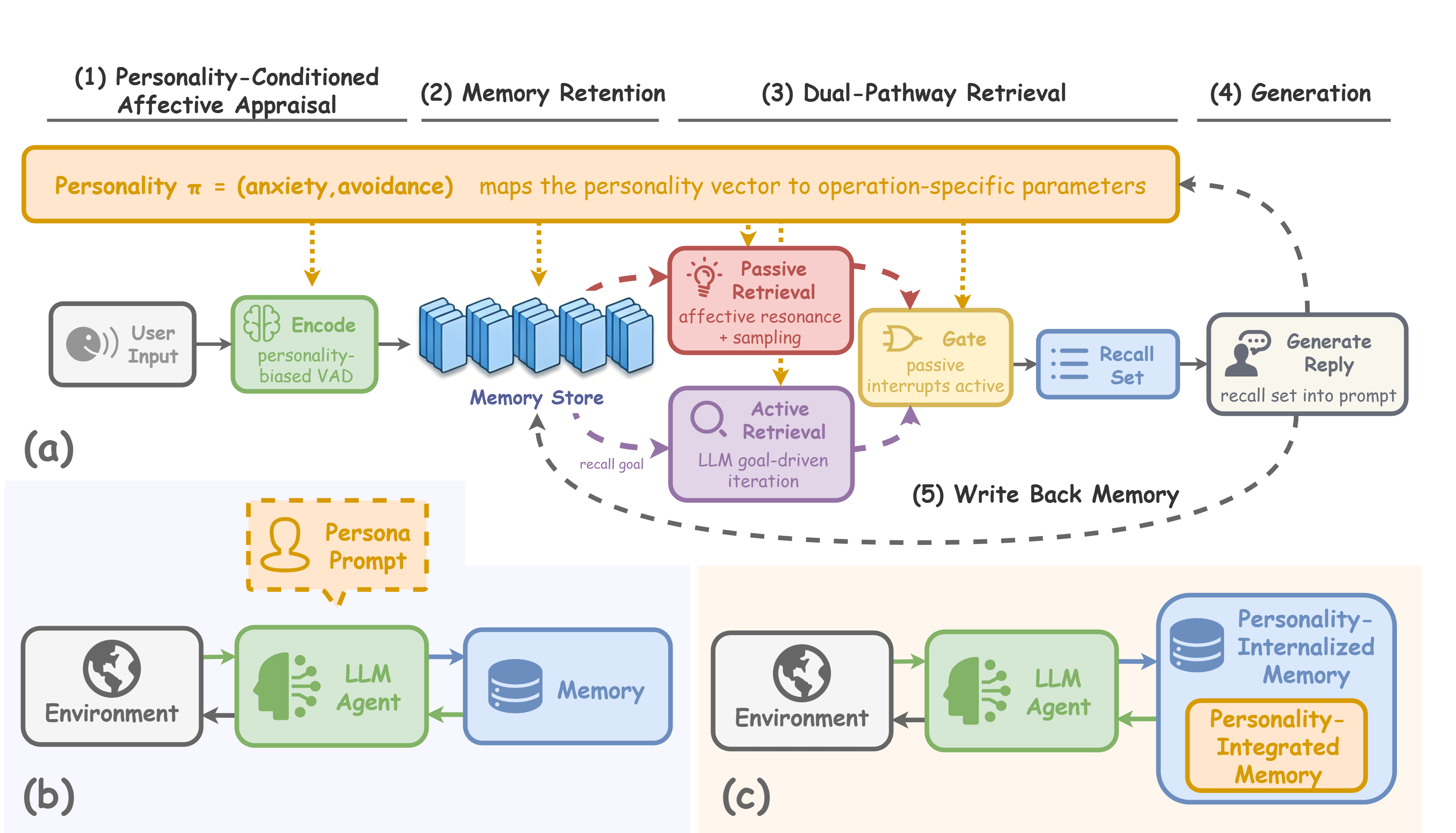}
  \caption{\textbf{Design Principle and Architectural Comparison of PersMem.}
  \textbf{(a)} A simplified view of PersMem. In the attachment instantiation, a fixed two-dimensional trait vector parameterises affective appraisal, memory retention, passive affect-driven memory retrieval, and the redirection of active goal-driven memory retrieval, while remaining hidden from the active controller and response model.
  \textbf{(b)} A common design specifies the assigned profile through a pre-defined personality in the system prompt while retrieving memories through a separate process.
  \textbf{(c)} PersMem instead uses one personality representation to control multiple memory operations before response generation. Fig.~\ref{fig:method} presents the complete processing and control flow.}
  \label{fig:overview}
\end{figure*}

Recently, role-playing agents built on large language models (LLMs) have been widely used in emotional companionship, clinical communication training, education, and interactive narrative games \cite{chen2024persona,tseng2024twotales}. These applications often require agents to maintain assigned profiles defined by pre-defined personality traits across extended human-agent interactions \cite{wang2024incharacter,tu2024charactereval,samuel2025personagym}. Such personality traits influence not only how an agent phrases its replies, but also how it retrieves and uses prior experiences, i.e., memories. This is evidenced by the research on mood-congruent memory,  which suggests that personality traits and affective states are associated with which autobiographical memories become accessible under the same cue \cite{bower1981mood,rusting1998personality}.

Most existing agent architectures define the pre-defined personality in the system prompt while managing stored memories through a separate process. In these agents, the memory module stores previous interactions, ranks candidate memories according to their semantic similarities with the current user input, and provides the selected memories to the response model. The personality description in the system prompt then guides how the response model interprets these memories and generates its reply, but does not directly control the preceding memory-processing operations or guide the memory module to retrieve memories according to the assigned personality \cite{park2023generative,zhong2024memorybank}. For example, when the user input describes a pleasant event, a personality-independent memory module may still prioritise positive memories of similar events and provide the response model with memory context that favours a cheerful reply rather than a negative reply consistent with the pre-defined personality.

In addition, personality specification in existing agent architectures remains dependent on the system prompt. When the interaction context becomes longer, or later instructions conflict with the system prompt, the influence of the personality description on reply generation may weaken \cite{kovac2024stick,abdulhai2025consistently,ding2026contextecho}. Previous research also shows that agent behaviours emerging during interactions may depart from the pre-defined identity \cite{zhang2026beyond}. Recent methods have started to use personality or affective information to condition parts of the agent pipeline \cite{xu2024character,huang2024emotionalrag}. However, these methods usually introduce such information at only one stage and do not use a unified and stable personality representation to control the entire memory-processing pipeline. Therefore, a new memory architecture is needed to allow the pre-defined personality to directly control memory processing and make its effects inspectable before reply generation.

To address these limitations, we propose Personality-Integrated Memory (PersMem), a novel autobiographical memory architecture that maps each pre-defined personality to operation-specific parameters controlling four personality-dependent memory operations: (i) affective appraisal, which assigns a personality-conditioned valence-arousal-dominance representation to the current user input; (ii) memory retention, which controls how previously stored memories remain accessible over time; (iii) passive affect-driven memory retrieval, which selects memories according to their semantic and affective relevance to the current user input; and (iv) active goal-driven memory retrieval, which searches for memories relevant to an explicit recall goal and can be redirected by passively retrieved memories. We instantiate this design in attachment and Big Five trait spaces using different theory-specific mappings while retaining the same memory-stage interfaces. Neither the active controller nor the response model receives the trait vector or an explicit persona instruction. As illustrated in Fig.~\ref{fig:overview}, PersMem allows the pre-defined personality to influence memory processing before response generation and exposes recalled-content statistics and redirection traces for analysis. Our main contributions are summarised as follows:
\begin{itemize}
\item We propose PersMem, a novel autobiographical memory architecture that represents a pre-defined personality as a fixed low-dimensional trait vector and maps this representation to operation-specific parameters controlling affective appraisal, memory retention, passive affect-driven memory retrieval, and active goal-driven memory retrieval, without providing the vector to the active controller or response model.

\item We introduce a personality-conditioned redirection mechanism that allows a memory proposed by passive affect-driven memory retrieval to redirect the next step of active goal-driven memory retrieval.

\item We evaluate personality-dependent memory-processing patterns and dialogue differences in attachment and Big Five settings, including under adversarial positive framing, and further assess character fidelity and narrative quality on the CoSER role-playing benchmark.
\end{itemize}

\section{Related Work}

\textbf{Memory Systems for LLM Agents.}
Memory systems for LLM agents commonly retrieve stored events according to semantic relevance, recency, or importance. Generative Agents combines these signals, while MemoryBank further introduces forgetting and reinforcement \cite{park2023generative,zhong2024memorybank}. Other systems manage long-term context through paging, consolidation, hierarchical storage, or layered character memory \cite{packer2023memgpt,chhikara2025mem0,kang2025memoryos,wang2026memorydriven,tang2026reveriemem}. These approaches improve memory storage and retrieval, but typically do not integrate a fixed personality representation across multiple memory operations.

\textbf{Personality-Aware Role-Playing Agents.}
Existing role-playing agents commonly represent personality through written personas or character data and evaluate whether generated behaviour remains faithful to the target character \cite{shao2023characterllm,zhou2024characterglm,wang2024incharacter,tu2024charactereval,samuel2025personagym}. Personality can also be elicited through Big Five questionnaires or generated text \cite{jiang2024personallm,serapio2023personality}, although prompted or questionnaire-based traits do not always match generated behaviour across contexts \cite{kovac2024stick,abdulhai2025consistently,ding2026contextecho,han2026personalityillusion,song2025mischaracterize}. Several methods connect persona or affect more directly to memory use: LD-Agent combines persona and event memories during response generation, CHARMAP performs persona-based retrieval, and Emotional RAG jointly considers semantic and emotional similarity \cite{li2025ldagent,xu2024character,huang2024emotionalrag}. Other approaches steer personality by modifying internal model activations \cite{zou2023representation,chen2025persona}. In contrast, PersMem uses one fixed trait vector to condition affective appraisal, memory retention, passive recall, and the redirection of active memory search, making personality-conditioned memory behaviour observable before response generation.

\textbf{Psychological Foundations.}
Mood-congruent memory relates affective state and personality to the accessibility of emotional information, while the self-memory system describes autobiographical recall as jointly shaped by the self and current goals \cite{bower1981mood,rusting1998personality,conway2000construction,conway2005memory}. Research on overgeneral autobiographical memory further motivates the functional-avoidance component of PersMem \cite{williams2007autobiographical}. Attachment theory provides a two-dimensional trait space over anxiety and avoidance, whereas the Big Five provides a five-dimensional personality representation \cite{bartholomew1991attachment,brennan1998selfreport,mccrae1987validation,goldberg2006ipip}. These theories motivate the memory operations and the two personality representations instantiated in PersMem.

\section{Personality-Integrated Memory}

\begin{figure*}[t]
  \centering
  \includegraphics[width=\textwidth]{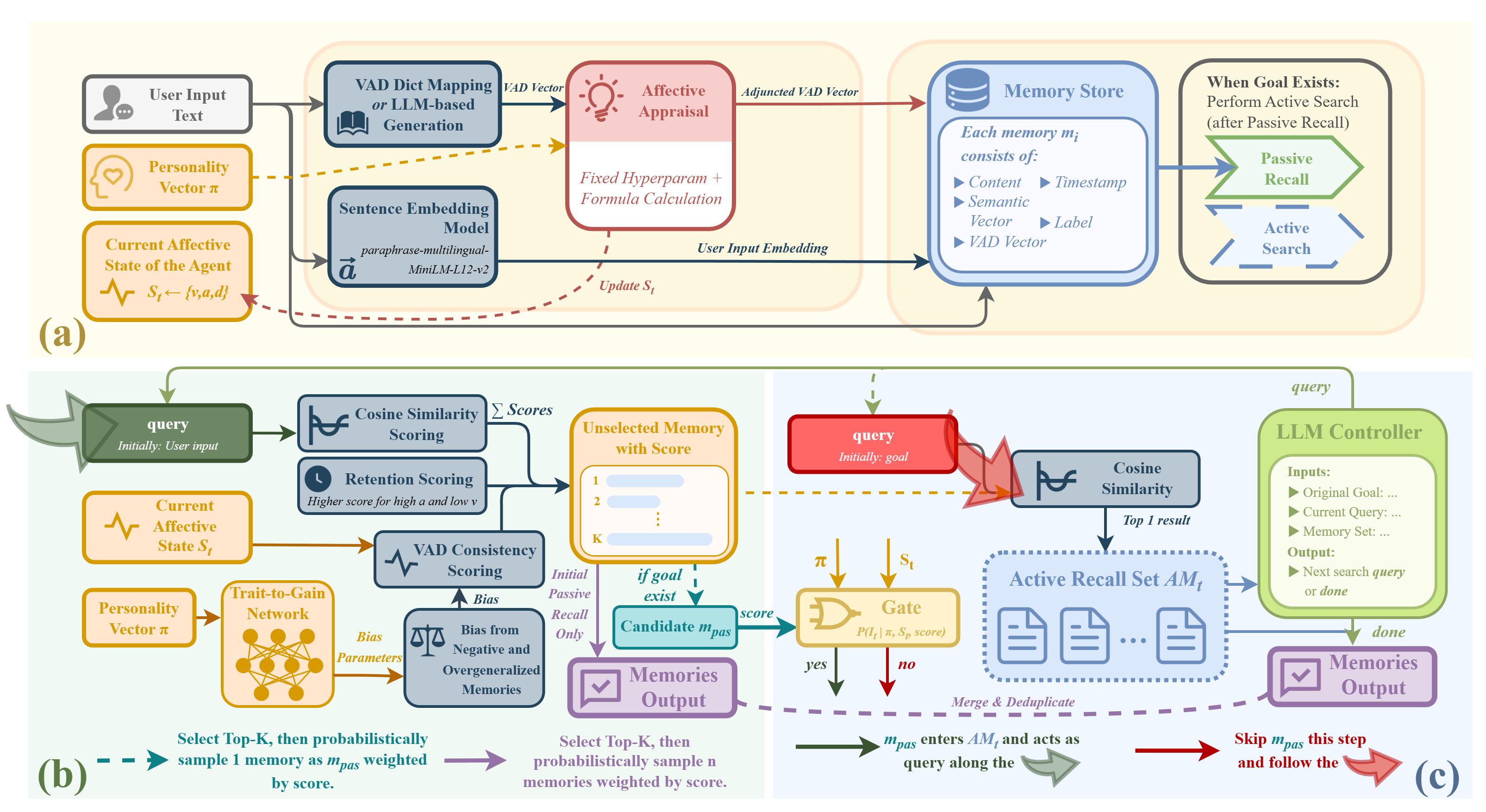}
  \caption{\textbf{Pipeline of PersMem.}
  \textbf{(a)} PersMem uses the pre-defined personality to annotate the affective state of the current user input, updates the agent's current state, and stores the completed exchange after reply generation.
  \textbf{(b)} Memory retention controls the accessibility of stored memories, while passive affect-driven memory retrieval selects an initial memory set and proposes one candidate during each active retrieval step.
  \textbf{(c)} Active goal-driven memory retrieval either adopts the candidate proposed by passive retrieval or continues semantic retrieval and query refinement. The initial passive output and active recall set are finally merged and deduplicated.}
  \label{fig:method}
\end{figure*}

\textbf{Overview.} As illustrated in Fig.~\ref{fig:method}, PersMem takes the current user input, the agent's current affective state, and the agent's pre-defined personality vector as inputs. \textbf{Affective appraisal} annotates the emotional state of the current user input and updates the agent's current affective state. \textbf{Memory retention} controls the forgetting of previously stored memories. \textbf{Passive affect-driven memory retrieval} scores memories according to semantic similarity, personality-conditioned affective consistency, and other factors, and selects relevant memories based on these scores. When an explicit recall goal is available, \textbf{active goal-driven memory retrieval} further searches for memories required by this goal, uses memories already retrieved by passive retrieval to guide subsequent retrieval steps, and outputs the final set of retrieved memories. Finally, the obtained memories are provided to the response model to generate the current reply, after which the completed interaction is stored as a new memory. During this process, neither the active retrieval controller nor the response model receives the personality representation. For methodological details, see Appendices B–D.

\textbf{Personality Representation.} PersMem represents the pre-defined personality using a fixed low-dimensional vector. The attachment setting uses $\pi=(\pi_{\mathrm{anx}},\pi_{\mathrm{avo}})\in[0,1]^2$, while the Big Five setting uses $\pi_{\mathrm{BF}}=(O,C,E,A,N)\in[0,1]^5$. Both settings use the same memory-processing pipeline but map their respective personality vectors to different parameters that control the following four operations.

\textbf{Affective Appraisal.} After receiving a user input, PersMem uses either an LLM-based annotator or an affective lexicon to convert it into a raw valence-arousal-dominance (VAD) affective vector $\tilde e=(v,a,d)$, where $v\in[-1,1]$ and $a,d\in[0,1]$. Meanwhile, a sentence embedding model converts the user input into a semantic embedding for subsequent use. PersMem then adjusts the raw VAD vector $\tilde e$ according to the pre-defined personality vector as $e=\mathcal A_{\pi}(\tilde e)$. In the attachment setting, higher anxiety strengthens negative valence, increases arousal for negative inputs, and reduces dominance, whereas higher avoidance dampens arousal and increases dominance. In the Big Five setting, neuroticism strengthens negative appraisal and arousal for negative inputs, extraversion and agreeableness strengthen positive appraisal, and lower agreeableness increases dominance.

During continuous interactions, PersMem maintains a dynamic VAD vector representing the agent's current affective state. After appraising the user input at turn $t$, PersMem updates this state as $s_t^{\mathrm{in}}=(1-\lambda)s_{t-1}^{\mathrm{out}}+\lambda e_t^{\mathrm{in}}$, where $s_t^{\mathrm{in}}$ denotes the affective state after appraising the user input at turn $t$, $s_{t-1}^{\mathrm{out}}$ is the affective state after the previous reply, $e_t^{\mathrm{in}}$ is the personality-adjusted VAD representation of the current input, and $\lambda\in[0,1]$ controls the contribution of the current input. After the agent's action module generates the reply, PersMem appraises the reply through the same process and computes the updated state $s_t^{\mathrm{out}}$. It then stores the completed interaction as a new memory.

\textbf{Memory Retention.} After receiving information from the previous module, including the VAD vector $e$ adjusted by the personality vector $\pi$, the memory store assigns a retention value to each currently stored memory. Specifically, the retention value of memory $m$ at time $t$ is defined as $\operatorname{ret}(m,t;\pi)=\exp(-(t-t_m)r(m;\pi))$, where $t_m$ denotes the time at which memory $m$ was stored, and $r(m;\pi)\geq 0$ is its personality-dependent decay rate. A larger $r(m;\pi)$ results in faster forgetting, whereas a smaller value keeps the memory accessible for longer. Memories with higher arousal decay more slowly. Attachment anxiety and Big Five neuroticism further slow the forgetting of negative memories, keeping them more accessible during subsequent retrieval. The resulting retention value is used as one component of the passive-retrieval score. 

\textbf{Passive Affect-Driven Memory Retrieval.} After receiving the outputs of the preceding modules, passive affect-driven memory retrieval scores each available memory. The score jointly considers the semantic similarity $\operatorname{sem}$ between the memory and the current query, the retention value $\operatorname{ret}$ computed by the preceding module, and the relation score $R$ between the memory's affective information and the agent's current affective state and pre-defined personality. In the attachment setting, the passive-retrieval score $S_{\mathrm{pas}}$ of a candidate memory is defined as $S_{\mathrm{pas}}=\alpha\,\operatorname{sem}+\beta_{\pi}R+\gamma\,\operatorname{ret}$, where $\beta_{\pi}$ is a weight modulated by the personality vector $\pi$ and controls the contribution of the affective relation score to the total score. The Big Five setting uses the same structure but adopts $\beta_{\mathrm{BF}}R_{\mathrm{BF}}$ as the personality-related affective retrieval term.

In the attachment setting, higher anxiety increases the scores of negative memories, while higher avoidance increases the scores of memories with higher dominance and further increases the scores of overgeneral memories when the current affective state is more negative. In the Big Five setting, higher neuroticism strengthens the tendency to retrieve negative memories, whereas higher extraversion and agreeableness strengthen the tendency to retrieve positive memories. Lower agreeableness increases the scores of memories with higher dominance, while lower conscientiousness together with higher neuroticism increases the scores of overgeneral memories. Openness does not directly add a new scoring term but instead modulates the temperature used in subsequent memory sampling. After scoring, PersMem forms a sampling distribution based on the candidate-memory scores and samples $n$ memories without replacement from the top-$K$ candidate memories to form the initial passive memory set. When an explicit recall goal is available, PersMem additionally performs active goal-driven memory retrieval. In this case, passive retrieval samples one additional candidate memory $m_{\mathrm{pas}}$ from the unused memories and then passes it into the active retrieval step.

\textbf{Active Goal-Driven Memory Retrieval.} When an explicit recall goal $g$ is available, PersMem additionally performs active goal-driven memory retrieval after passive retrieval and uses the goal as the initial search query, $q_0=g$. At the $t$-th retrieval step, passive retrieval first uses the current search query $q_t$ to sample one additional candidate memory $m_t^{\mathrm{pas}}$ from the unused memories. A personality-conditioned redirection gate then determines, with probability $I_t=\sigma(\lambda_I[S_{\mathrm{pas}}(m_t^{\mathrm{pas}})-\theta]+\nu a_t)$, whether to use this candidate memory to redirect the next retrieval step. Here, $S_{\mathrm{pas}}(m_t^{\mathrm{pas}})$ denotes the passive-retrieval score of the candidate memory $m_t^{\mathrm{pas}}$, $a_t$ denotes the arousal of the agent's current affective state, $\theta$ denotes the personality-conditioned firing threshold, and $\sigma$ denotes the Sigmoid function. The coefficient $\lambda_I$ controls the effect of the difference between the candidate-memory score and the firing threshold on the firing probability, while $\nu$ controls the effect of the current arousal. In the attachment setting, the firing threshold is defined as $\theta_\pi=\theta_0-\eta\pi_{\mathrm{anx}}$; in the Big Five setting, it is defined as $\theta_{\mathrm{BF}}=\theta_0-\eta N$. Here, $\theta_0$ is the base firing threshold, $\eta$ controls the strength of personality modulation on the threshold, and $\pi_{\mathrm{anx}}$ and $N$ denote attachment anxiety and neuroticism, respectively. Therefore, higher attachment anxiety or neuroticism lowers the firing threshold and makes the redirection gate more likely to fire.

When the gate fires, PersMem adds the candidate memory $m_t^{\mathrm{pas}}$ to the active recall set and uses the beginning of its text as $q_{t+1}$ to restart active retrieval. When the gate does not fire, $m_t^{\mathrm{pas}}$ is not adopted at the current step. PersMem instead retrieves the unused memory that is most semantically similar to the current query and adds it to the active recall set. An LLM controller then receives the original recall goal, the current search query, and excerpts from the memories accumulated during active retrieval, and outputs either a revised search query for the next step or \texttt{DONE}. Active retrieval stops when the controller outputs \texttt{DONE} or an empty query, when no unused memory remains, or when the maximum number of retrieval steps is reached. Finally, PersMem merges and deduplicates the initial passive memory set and the active recall set, and provides the resulting memory set together with the current user input to the response model. The response model receives neither the personality vector nor an explicit persona instruction; the complete response-prompt configurations are provided in the corresponding experimental settings.

\section{Experiments}

We evaluate PersMem at both the memory-processing and final-dialogue levels. Since PersMem is designed to make personality affect memory processing before response generation, our primary evaluation examines the resulting memory-processing behaviour rather than relying only on final dialogues. Experiment~1 tests whether different attachment personalities produce distinguishable memory-processing patterns and whether these differences weaken when personality-dependent operations are removed. Experiment~2 examines whether continuous Big Five personality traits are associated with the characteristics of recalled memories and whether these differences remain visible in the generated dialogues. Experiment~3 evaluates PersMem on CoSER to assess its performance in character-based role-playing conversations. Additional evaluations examine robustness under adversarial framing and performance on a general memory question-answering task.

Existing methods differ from PersMem in both their memory-processing mechanisms and evaluation targets, and no established benchmark directly evaluates personality-dependent memory processing. We therefore focus on quantitative and interpretability analyses that test whether PersMem changes the intended memory operations, whether these changes affect downstream dialogues, and whether the system remains robust and useful on broader memory tasks. In Experiments~1 and~2, neither the agent's action module nor PersMem's active retrieval controller receives the personality vector or an explicit persona instruction.

\subsection{Experiment 1: Attachment Personality Effects on Memory Processing}

\textbf{Setup.} We evaluate four attachment personalities: secure, anxious, avoidant, and fearful. We use separate memory sets for parameter calibration, classifier development, and final testing, containing 63, 92, and 165 memories, respectively. No memory is shared across these sets, and the development and test cues do not overlap. Recall evaluation uses autobiographical memories with VAD annotations, while overgeneral-memory (OGM) evaluation uses emotional cues and memories with specificity labels.

The classifier is trained on the development set and fixed before testing. It predicts the assigned attachment personality using only recalled-content statistics and gate-redirection traces. It does not receive the personality vector, internal resonance or retention scores, or gate probabilities. GPT-4o generates the responses, and Qwen3.5-9B serves as an independent evaluator. Recall and OGM analyses include all four personalities, while active-retrieval and response analyses compare the secure, anxious, and avoidant personalities. Appendix~E provides the complete data construction, features, and preprocessing details.

\textbf{Attachment-personality separability.}
Table~\ref{tab:attachment-classification} reports the primary classification results on the held-out test set. Full PersMem achieves 48.1\% accuracy, compared with the 25.0\% chance level for four-way classification. Accuracy decreases to 40.0\% when the personality-conditioned gate is replaced with a neutral gate and falls to chance when the gate is disabled. A classifier using only recalled-memory text embeddings also remains at chance, while a linear trait-aware reranker achieves 30.0\%. These results show that PersMem produces distinguishable attachment-related memory-processing patterns that cannot be explained by recalled-memory text embeddings alone. Complete-trace diagnostics and comparisons with retrieval baselines are reported in Appendix~E because they include internal scores and use different data splits.

\begin{table}[t]
\centering
\footnotesize
\setlength{\tabcolsep}{3pt}
\renewcommand{\arraystretch}{1.08}
\caption{Attachment profile classification. Held-out evaluation excludes internal personality-dependent scores; complete-trace diagnostics include resonance and retention scores. Primary and matched complete-trace results use different splits.}
\label{tab:attachment-classification}
\begin{tabularx}{\columnwidth}{
  @{}
  >{\raggedright\arraybackslash}X
  >{\centering\arraybackslash}p{0.15\columnwidth}
  >{\raggedright\arraybackslash}p{0.31\columnwidth}
  @{}
}
\toprule
Condition & Accuracy & 95\% CI / $p$-value \\
\midrule

\multicolumn{3}{@{}l}{\textit{Held-out classification}} \\
Full PersMem
& 48.1\%
& [40.0\%, 56.3\%]; $p=0.0002$ \\

Neutral gate
& 40.0\%
& -- \\

Gate off
& 25.0\%
& -- \\

Text embedding only
& 25.0\%
& -- \\

Linear trait-aware reranker
& 30.0\%
& $p=0.055$ \\

\midrule
\multicolumn{3}{@{}l}{\textit{Complete trace, primary split}} \\
PersMem
& 75.0\%
& [66.6\%, 81.9\%] \\

\midrule
\multicolumn{3}{@{}l}{\textit{Complete trace, matched split}} \\
PersMem
& 72.5\%
& -- \\

Semantic RAG
& 28.3\%
& -- \\

Generative-Agents-style scoring
& 25.8\%
& -- \\

No-trait control
& 26.7\%
& -- \\

\bottomrule
\end{tabularx}
\end{table}

\textbf{Contributions of the four memory operations.}
Table~\ref{tab:attachment-checks} reports the average marginal effect of each personality-dependent operation on held-out classification. Passive affect-driven memory retrieval has the largest effect $(+0.130)$, followed by personality-conditioned redirection within active retrieval $(+0.085)$. Affective appraisal $(+0.005)$ and memory retention $(+0.013)$ have smaller individual effects. These values describe average marginal contributions and do not measure interactions among the four operations.

Replacing the primary VAD annotations with NRC-VAD \cite{mohammad2025nrcvad} produces a similar above-chance classification accuracy of 45.0\%. An independent language model also reproduces 99.4\% of the original specificity labels and yields the same directional OGM effects. These results reduce dependence on a single annotation procedure. Further annotation results are provided in Appendix~E.

\begin{table}[t]
\centering
\small
\setlength{\tabcolsep}{3pt}
\renewcommand{\arraystretch}{1.08}
\caption{Component effects and annotation checks.}
\label{tab:attachment-checks}
\begin{tabularx}{\columnwidth}{
  @{}
  >{\raggedright\arraybackslash}X
  >{\centering\arraybackslash}p{0.15\columnwidth}
  >{\centering\arraybackslash}p{0.23\columnwidth}
  >{\centering\arraybackslash}p{0.16\columnwidth}
  @{}
}
\toprule
Analysis & Estimate & CI or statistic & $p$-value \\
\midrule
Affective-appraisal effect
& $+0.005$ & -- & -- \\

Memory-retention effect
& $+0.013$ & -- & -- \\

Passive-retrieval effect
& $+0.130$ & -- & -- \\

Active-retrieval redirection effect
& $+0.085$ & -- & -- \\
\midrule
Primary VAD classification
& 48.1\% & [40.0\%, 56.3\%] & $0.0002$ \\

NRC-VAD classification
& 45.0\% & [38.8\%, 51.2\%] & $0.0002$ \\

Specificity-label agreement
& 99.4\% & $\kappa=0.988$ & -- \\

Fearful specificity effect, neutral
& $+0.0422$ & -- & $1\times10^{-5}$ \\

Fearful specificity effect, distress
& $+0.0770$ & -- & $1\times10^{-5}$ \\
\bottomrule
\end{tabularx}
\end{table}

\textbf{Personality-dependent retrieval and response behaviour.}
Table~\ref{tab:attachment-profile-results} summarises the main behavioural differences. Avoidant and fearful personalities retrieve more overgeneral memories under distress, consistent with the implemented avoidance- and distress-dependent OGM mapping. Among the personalities included in active retrieval, the anxious personality has the highest proportion of redirections toward negative memories. This indicates that its redirection gate more frequently introduces negative memories into the active recall set.

Attachment-related differences are also detectable in the generated responses: the response-profile classifier achieves $16/30=53.3\%$ accuracy, compared with the 33.3\% chance level. However, this result is descriptive because the ten responses generated for each personality form a continuous sequence rather than independent observations. Experiment~2 therefore provides the primary response-level evaluation. Detailed profile-level statistics are reported in Appendix~E.

\begin{table}[t]
\centering
\footnotesize
\setlength{\tabcolsep}{1.5pt}
\renewcommand{\arraystretch}{1.08}
\caption{Profile-conditioned retrieval and response outcomes.
Dashes indicate profiles excluded from an analysis.}
\label{tab:attachment-profile-results}
\begin{tabularx}{\columnwidth}{
  @{}
  >{\raggedright\arraybackslash}X
  *{4}{>{\centering\arraybackslash}p{0.125\columnwidth}}
  @{}
}
\toprule
Measure & Sec. & Anx. & Avo. & Fear. \\
\midrule
Neutral OGM
& -- & -- & 0.28 & 0.47 \\

Distress OGM
& -- & -- & 0.37 & 0.73 \\

OGM lift over no-trait
& -- & -- & $+0.07$ & $+0.27$ \\

Independent-label OGM
difference from secure, distress
& 0 & $+0.027$ & $+0.039$ & $+0.067$ \\

Neutral mean recalled valence
& -- & -- & $-0.04$ & -- \\
\midrule
Negative-redirection proportion
& 0.59 & 0.63 & 0.51 & -- \\

Response negativity score
& 3.0 & 5.0 & 2.0 & -- \\
\midrule
\multicolumn{5}{@{}l@{}}{
Response-profile classification: $16/30=53.3\%$; chance $=33.3\%$
} \\
\bottomrule
\end{tabularx}
\end{table}

\subsection{Experiment 2: Big Five Personality Effects on Recall and Dialogue}

\textbf{Setup.}
Each sampled profile is represented by a fixed Big Five personality vector $\pi_{\mathrm{BF}}=(O,C,E,A,N)$. We sample 50 profiles from the OpenPsychometrics dataset based on the IPIP Big-Five Factor Markers \cite{goldberg1992development}, while retaining all five trait values of each respondent. PersMem uses the same memory-processing pipeline as in Experiment~1, with the Big Five-dependent mappings described in Appendix~D.

Each profile completes the same six-round dialogue and retrieves three memories per round. The memory corpus contains 320 autobiographical events with VAD and specificity labels, evenly divided between specific and overgeneral memories. Qwen3.5-9B generates the dialogues, while Llama-3.1-8B serves as a blind cross-model evaluator. Recall analyses use all 50 profiles, whereas dialogue evaluation compares profiles with high and low values of each trait.

We compare the full PersMem system with three controls. The no-trait control removes all Big Five-dependent adjustments while retaining the same memory-retrieval pipeline. The random-memory control uniformly samples three memories in each round, and the no-memory control provides no retrieved memories. All conditions use the same profiles, prompts, memory corpus, number of dialogue rounds, and response model. Complete profile construction and experimental settings are provided in Appendix~F.

\textbf{Personality-related recall patterns.}
We first examine whether continuous Big Five traits are associated with the characteristics of recalled memories. For each profile, PersMem retrieves 18 memories across six shared prompts. As shown in Table~\ref{tab:ocean-recall}, higher neuroticism is associated with more negative recalled valence $(r=-0.703)$, whereas higher extraversion and agreeableness are associated with more positive recalled valence $(r=0.776$ and $r=0.607)$. Higher conscientiousness is also associated with fewer overgeneral memories under distress $(r=-0.710)$.

Under the no-trait control, all profiles receive the same recalled-valence result, so the corresponding correlations cannot be computed. Its association between conscientiousness and overgeneral-memory retrieval is also substantially weaker $(r=-0.180)$. These results show that the implemented Big Five mappings produce distinguishable recall patterns across continuous personality profiles. Because all five traits vary together within each sampled profile, these results describe profile-level associations rather than the isolated effect of any single trait.

\begin{table}[t]
\centering
\small
\setlength{\tabcolsep}{3pt}
\renewcommand{\arraystretch}{1.08}
\caption{Descriptive Big Five recall associations over 50 profiles. Dashes indicate undefined valence correlations because the no-trait control produces zero variance in recalled valence.}
\label{tab:ocean-recall}
\begin{tabularx}{\columnwidth}{
  @{}
  >{\raggedright\arraybackslash}X
  >{\raggedright\arraybackslash}p{0.23\columnwidth}
  >{\centering\arraybackslash}p{0.17\columnwidth}
  >{\centering\arraybackslash}p{0.19\columnwidth}
  @{}
}
\toprule
Trait & Outcome & Full system & No-trait control \\
\midrule
Neuroticism
& Valence & $-0.703$ & -- \\

Extraversion
& Valence & $+0.776$ & -- \\

Agreeableness
& Valence & $+0.607$ & -- \\

Conscientiousness
& OGM & $-0.710$ & $-0.180$ \\
\bottomrule
\end{tabularx}
\end{table}

\textbf{Personality-related dialogue differences.}
We next examine whether these personality-related memory differences are reflected in the generated dialogues. For each Big Five trait, we select the 12 profiles with the highest values and the 12 profiles with the lowest values, and pair them by rank. This produces 60 high--low dialogue pairs across the five traits. Each pair is evaluated in both presentation orders, resulting in 120 evaluation calls per condition.

The evaluator receives two six-round dialogues and a description of the target trait, but does not receive their trait values, retrieved memories, condition labels, or generation prompts. It is asked to identify which dialogue was generated for the profile with the higher assigned trait value.

As shown in Table~\ref{tab:ocean-dialogue}, the full PersMem system achieves 67.5\% accuracy. The random-memory control reaches 60.8\%, while the no-trait and no-memory controls reach 54.2\% and 53.3\%, respectively. These results indicate that personality-dependent memory processing produces trait-related differences that remain detectable in the final dialogues. However, the 6.7-percentage-point difference between PersMem and the random-memory control is descriptive because each dialogue pair is evaluated twice in different presentation orders. Additional generator and evaluator configurations are reported in Appendix~F.

\begin{table}[t]
\centering
\small
\setlength{\tabcolsep}{4pt}
\caption{Blind Big Five dialogue comparison. Each of 60 high--low pairs is evaluated in two presentation orders.}
\label{tab:ocean-dialogue}
\begin{tabular}{@{}lcc@{}}
\toprule
Condition & Correct calls & Accuracy \\
\midrule
Full system
& \textbf{81/120} & \textbf{67.5\%} \\

Random-memory control
& 73/120 & 60.8\% \\

No-trait control
& 65/120 & 54.2\% \\

No-memory control
& 64/120 & 53.3\% \\
\bottomrule
\end{tabular}
\end{table}

\subsection{Experiment 3: Role-Playing Evaluation on CoSER}

\textbf{Setup.}
We evaluate PersMem on CoSER~\cite{wang2025coser}, using \texttt{DeepSeek-v4-flash} as the evaluator. We report four dimensions: Storyline Consistency (SC), Anthropomorphism (AN), Character Fidelity (CF), and Storyline Quality (SQ), together with their average. The complete evaluation configuration, supplementary metrics, and sources of the published reference results are provided in Appendix~I.

\textbf{Results.}
Table~\ref{tab:coser-results} presents the PersMem results alongside published reference results. PersMem achieves an average score of 66.13, with 59.83 for Storyline Consistency, 51.00 for Anthropomorphism, 69.33 for Character Fidelity, and 84.33 for Storyline Quality. This evaluation complements the attachment and Big Five experiments by assessing character portrayal and narrative quality in an external role-playing setting.

\begin{table}[t]
\centering
\footnotesize
\setlength{\tabcolsep}{2.5pt}
\renewcommand{\arraystretch}{1.08}
\caption{Role-playing results on CoSER. Published reference scores retain their source studies' evaluation settings; our PersMem results use DeepSeek. SC, AN, CF, and SQ denote Storyline Consistency, Anthropomorphism, Character Fidelity, and Storyline Quality, respectively. Conv. denotes conversation retrieval.}
\label{tab:coser-results}
\begin{tabularx}{\columnwidth}{
@{}
>{\raggedright\arraybackslash}X
rrrrr
@{}
}
\toprule
Method & SC & AN & CF & SQ & Avg. \\
\midrule

\multicolumn{6}{@{}l@{}}{\textit{CoSER}~\cite{wang2025coser}} \\
GPT-4o
& 61.59 & 48.93 & 48.95 & 80.33 & 59.95 \\
CoSER-70B + Conv.
& 64.59 & 53.79 & 54.86 & 77.28 & 62.63 \\

\midrule
\multicolumn{6}{@{}l@{}}{\textit{CogDual}~\cite{liu2025cogdual}} \\
LLaMA3.1-8B + CogDual-RL
& 60.10 & 45.89 & 48.82 & 73.08 & 56.97 \\

\midrule
\multicolumn{6}{@{}l@{}}{\textit{HER}~\cite{du2026her}} \\
Claude-4.5-Opus
& 63.74 & 64.28 & 58.45 & 63.24 & 62.43 \\
Gemini-3-Pro
& 65.95 & 60.42 & 58.34 & 62.49 & 61.80 \\
GPT-5.1
& 64.95 & 53.99 & 60.13 & 65.35 & 61.10 \\

\midrule
\multicolumn{6}{@{}l@{}}{\textit{Our evaluation}} \\
\textbf{PersMem}
& 59.83 & 51.00 & 69.33 & 84.33 & 66.13 \\

\bottomrule
\end{tabularx}
\end{table}

\subsection{Supplementary Evaluations}
We further examine the fitted personality mappings, robustness to adversarial instructions, and performance on a general memory question-answering task. The fitted trait-to-gain mapping obtains a lower calibration error than a fixed-gain reference $(0.084$ versus $0.482)$. When the agent is instructed to discuss only positive content, PersMem retains 89\% of its original neuroticism-related retrieval difference, compared with 58\% for a prompt-persona baseline.

On the LoCoMo benchmark \cite{maharana2024evaluating}, PersMem improves mean answer F1 over Generative-Agents-style scoring \cite{park2023generative} by 0.0169, with a 95\% confidence interval of $[0.0027,0.0310]$. It also increases evidence recall@5 from 0.341 to 0.387. The two methods achieve similar answer F1 when the required evidence is successfully retrieved, suggesting that the improvement mainly comes from retrieving the relevant evidence more often. Semantic RAG obtains the highest evidence recall@5 of 0.444. Complete calibration, adversarial-framing, cross-model, and LoCoMo settings are provided in Appendices~D, F, and~H.

\begin{table}[t]
\centering
\footnotesize
\setlength{\tabcolsep}{3pt}
\renewcommand{\arraystretch}{1.08}
\caption{Calibration and adversarial-framing results.}
\label{tab:ocean-calibration-adversarial}
\begin{tabularx}{\columnwidth}{
  @{}
  >{\raggedright\arraybackslash}p{0.29\columnwidth}
  >{\raggedright\arraybackslash}X
  >{\centering\arraybackslash}p{0.17\columnwidth}
  @{}
}
\toprule
Evaluation & Condition & Result \\
\midrule

\multirow[t]{2}{0.29\columnwidth}{Calibration error}
& Fitted gain map
& 0.084 \\

& Fixed-gain reference
& 0.482 \\

\midrule

\multirow[t]{2}{0.29\columnwidth}{Adversarial retention}
& PersMem
& 89\% \\

& Prompt-persona baseline
& 58\% \\

\bottomrule
\end{tabularx}
\end{table}

\begin{table}[t]
\centering
\footnotesize
\setlength{\tabcolsep}{3pt}
\renewcommand{\arraystretch}{1.08}
\caption{LoCoMo retrieval and answer results.}
\label{tab:locomo-results}
\begin{tabularx}{\columnwidth}{
  @{}
  >{\raggedright\arraybackslash}X
  >{\centering\arraybackslash}p{0.25\columnwidth}
  >{\centering\arraybackslash}p{0.25\columnwidth}
  @{}
}
\toprule
Method
& Evidence recall@5
& F1 given evidence hit \\
\midrule
PersMem
& 0.387
& 0.547 \\

Generative-Agents-style scoring
& 0.341
& 0.540 \\

Semantic RAG
& 0.444
& 0.542 \\
\bottomrule
\end{tabularx}
\end{table}

\section{Conclusion}

In this paper, we presented PersMem, a memory architecture that maps a fixed low-dimensional personality vector to affective appraisal, memory retention, passive affect-driven memory retrieval, and personality-conditioned redirection within active goal-driven memory retrieval. Neither the active retrieval controller nor the response model receives the personality vector or an explicit persona instruction. Experiments with attachment and Big Five personalities show that PersMem produces distinguishable personality-dependent memory-processing patterns, and that some of these differences remain visible in the generated dialogues. The CoSER evaluation extends this assessment to character fidelity and narrative quality in role-playing conversations, where PersMem achieves an average score of 66.13. PersMem therefore provides a direct and inspectable way to integrate a pre-defined personality into memory processing before response generation. It is worth noting that the personality-dependent mappings implemented in PersMem are design choices and should not be interpreted as validated computational models of human personality. Future work should validate the resulting memory and dialogue behaviour through human judgements, examine longer interactions, and evaluate PersMem on broader personality and memory settings.

\bibliography{aaai2027}

@article{chen2024persona,
  title={From Persona to Personalization: A Survey on Role-Playing Language Agents},
  author={Chen, Jiangjie and Wang, Xintao and Xu, Rui and Yuan, Siyu and Zhang, Yikai and Shi, Wei and Xie, Jian and Li, Shuang and Yang, Ruihan and Zhu, Tinghui and Chen, Aili and Li, Nianqi and Chen, Lida and Hu, Caiyu and Wu, Siye and Ren, Scott and Fu, Ziquan and Xiao, Yanghua},
  journal={Transactions on Machine Learning Research (TMLR)},
  year={2024},
  note={arXiv:2404.18231}
}

@inproceedings{tseng2024twotales,
  title={Two tales of persona in llms: A survey of role-playing and personalization},
  author={Tseng, Yu-Min and Huang, Yu-Chao and Hsiao, Teng-Yun and Chen, Wei-Lin and Huang, Chao-Wei and Meng, Yu and Chen, Yun-Nung},
  booktitle={Findings of the Association for Computational Linguistics: EMNLP 2024},
  pages={16612--16631},
  year={2024}
}

@inproceedings{wang2024incharacter,
  title={Incharacter: Evaluating personality fidelity in role-playing agents through psychological interviews},
  author={Wang, Xintao and Xiao, Yunze and Huang, Jen-tse and Yuan, Siyu and Xu, Rui and Guo, Haoran and Tu, Quan and Fei, Yaying and Leng, Ziang and Wang, Wei and others},
  booktitle={Proceedings of the 62nd annual meeting of the association for computational linguistics (volume 1: Long papers)},
  pages={1840--1873},
  year={2024}
}

@inproceedings{tu2024charactereval,
  title={Charactereval: A chinese benchmark for role-playing conversational agent evaluation},
  author={Tu, Quan and Fan, Shilong and Tian, Zihang and Shen, Tianhao and Shang, Shuo and Gao, Xin and Yan, Rui},
  booktitle={Proceedings of the 62nd Annual Meeting of the Association for Computational Linguistics (Volume 1: Long Papers)},
  pages={11836--11850},
  year={2024}
}

@inproceedings{samuel2025personagym,
  title={PersonaGym: Evaluating Persona Agents and LLMs},
  author={Samuel, Vinay and Zou, Henry Peng and Zhou, Yue and Chaudhari, Shreyas and Kalyan, Ashwin and Rajpurohit, Tanmay and Deshpande, Ameet and Narasimhan, Karthik and Murahari, Vishvak},
  booktitle={Findings of the Association for Computational Linguistics: EMNLP 2025},
  pages={6999--7022},
  year={2025}
}

@inproceedings{shao2023characterllm,
  title={Character-llm: A trainable agent for role-playing},
  author={Shao, Yunfan and Li, Linyang and Dai, Junqi and Qiu, Xipeng},
  booktitle={Proceedings of the 2023 Conference on Empirical Methods in Natural Language Processing},
  pages={13153--13187},
  year={2023}
}

@inproceedings{zhou2024characterglm,
  title={CharacterGLM: Customizing social characters with large language models},
  author={Zhou, Jinfeng and Chen, Zhuang and Wan, Dazhen and Wen, Bosi and Song, Yi and Yu, Jifan and Huang, Yongkang and Ke, Pei and Bi, Guanqun and Peng, Libiao and others},
  booktitle={Proceedings of the 2024 conference on empirical methods in natural language processing: Industry track},
  pages={1457--1476},
  year={2024}
}

@inproceedings{park2023generative,
  title={Generative agents: Interactive simulacra of human behavior},
  author={Park, Joon Sung and O'Brien, Joseph and Cai, Carrie Jun and Morris, Meredith Ringel and Liang, Percy and Bernstein, Michael S},
  booktitle={Proceedings of the 36th annual acm symposium on user interface software and technology},
  pages={1--22},
  year={2023}
}

@inproceedings{zhong2024memorybank,
  title={Memorybank: Enhancing large language models with long-term memory},
  author={Zhong, Wanjun and Guo, Lianghong and Gao, Qiqi and Ye, He and Wang, Yanlin},
  booktitle={Proceedings of the AAAI conference on artificial intelligence},
  volume={38},
  number={17},
  pages={19724--19731},
  year={2024}
}

@article{kovac2024stick,
title={Stick to your role! stability of personal values expressed in large language models},
  author={Kova{\v{c}}, Grgur and Portelas, R{\'e}my and Sawayama, Masataka and Dominey, Peter Ford and Oudeyer, Pierre-Yves},
  journal={Plos one},
  volume={19},
  number={8},
  pages={e0309114},
  year={2024},
  publisher={Public Library of Science San Francisco, CA USA}
}

@inproceedings{abdulhai2025consistently,
  title     = {Consistently Simulating Human Personas with Multi-Turn Reinforcement Learning},
  author    = {Abdulhai, Marwa and Cheng, Ryan and Clay, Donovan
               and Althoff, Tim and Levine, Sergey and Jaques, Natasha},
  booktitle = {Advances in Neural Information Processing Systems},
  volume    = {38},
  pages     = {52920--52957},
  year      = {2025},
  publisher = {Curran Associates, Inc.},
  url       = {https://proceedings.neurips.cc/paper_files/paper/2025/hash/4c91443877f8388d8190c938ac5a4d4d-Abstract-Conference.html}
}

@article{ding2026contextecho,
  title   = {{ContextEcho}: A Benchmark for Persona Drift in Long Agentic-Coding Sessions},
  author  = {Ding, Xianzhong and Yu, Yangyang and Liu, Changwei
             and Zhao, Bill},
  journal = {arXiv preprint arXiv:2605.24279},
  year    = {2026},
  doi     = {10.48550/arXiv.2605.24279},
  url     = {https://arxiv.org/abs/2605.24279}
}

@article{bower1981mood,
  title={Mood and memory.},
  author={Bower, Gordon H},
  journal={American psychologist},
  volume={36},
  number={2},
  pages={129},
  year={1981},
  publisher={American Psychological Association}
}

@article{rusting1998personality,
  title={Personality, mood, and cognitive processing of emotional information: three conceptual frameworks.},
  author={Rusting, Cheryl L},
  journal={Psychological bulletin},
  volume={124},
  number={2},
  pages={165},
  year={1998},
  publisher={American Psychological Association}
}

@article{conway2000construction,
  title={The construction of autobiographical memories in the self-memory system.},
  author={Conway, Martin A and Pleydell-Pearce, Christopher W},
  journal={Psychological review},
  volume={107},
  number={2},
  pages={261},
  year={2000},
  publisher={American Psychological Association}
}

@article{conway2005memory,
  title={Memory and the self},
  author={Conway, Martin A},
  journal={Journal of memory and language},
  volume={53},
  number={4},
  pages={594--628},
  year={2005},
  publisher={Elsevier}
}

@article{bartholomew1991attachment,
  title={Attachment styles among young adults: a test of a four-category model.},
  author={Bartholomew, Kim and Horowitz, Leonard M},
  journal={Journal of personality and social psychology},
  volume={61},
  number={2},
  pages={226},
  year={1991},
  publisher={American Psychological Association}
}

@incollection{brennan1998selfreport,
  title     = {Self-Report Measurement of Adult Attachment: An Integrative Overview},
  author    = {Brennan, Kelly A. and Clark, Catherine L.
               and Shaver, Phillip R.},
  editor    = {Simpson, Jeffry A. and Rholes, W. Steven},
  booktitle = {Attachment Theory and Close Relationships},
  pages     = {46--76},
  publisher = {Guilford Press},
  address   = {New York, NY},
  year      = {1998}
}

@article{mccrae1987validation,
  title={Validation of the five-factor model of personality across instruments and observers.},
  author={McCrae, Robert R and Costa, Paul T},
  journal={Journal of personality and social psychology},
  volume={52},
  number={1},
  pages={81},
  year={1987},
  publisher={American Psychological Association}
}

@article{goldberg2006ipip,
  title={The international personality item pool and the future of public-domain personality measures},
  author={Goldberg, Lewis R and Johnson, John A and Eber, Herbert W and Hogan, Robert and Ashton, Michael C and Cloninger, C Robert and Gough, Harrison G},
  journal={Journal of Research in personality},
  volume={40},
  number={1},
  pages={84--96},
  year={2006},
  publisher={Elsevier}
}

@article{williams2007autobiographical,
  title={Autobiographical memory specificity and emotional disorder.},
  author={Williams, J Mark G and Barnhofer, Thorsten and Crane, Catherine and Herman, Dirk and Raes, Filip and Watkins, Ed and Dalgleish, Tim},
  journal={Psychological bulletin},
  volume={133},
  number={1},
  pages={122},
  year={2007},
  publisher={American Psychological Association}
}

@article{packer2023memgpt,
  title={MemGPT: towards LLMs as operating systems},
  author={Packer, Charles and Fang, Vivian and Patil, Shishir\_G and Lin, Kevin and Wooders, Sarah and Gonzalez, Joseph\_E},
  year={2023},
  publisher={ArXiv}
}

@inproceedings{kang2025memoryos,
  title={Memory os of ai agent},
  author={Kang, Jiazheng and Ji, Mingming and Zhao, Zhe and Bai, Ting},
  booktitle={Proceedings of the 2025 Conference on Empirical Methods in Natural Language Processing},
  pages={25972--25981},
  year={2025}
}

@inproceedings{li2025ldagent,
  title={Hello again! llm-powered personalized agent for long-term dialogue},
  author={Li, Hao and Yang, Chenghao and Zhang, An and Deng, Yang and Wang, Xiang and Chua, Tat-Seng},
  booktitle={Proceedings of the 2025 Conference of the Nations of the Americas Chapter of the Association for Computational Linguistics: Human Language Technologies (Volume 1: Long Papers)},
  pages={5259--5276},
  year={2025}
}

@article{xu2024character,
  title={Character is destiny: Can role-playing language agents make personadriven decisions},
  author={Xu, Rui and Wang, Xintao and Chen, Jiangjie and Yuan, Siyu and Yuan, Xinfeng and Liang, Jiaqing and Chen, Zulong and Dong, Xiaoqing and Xiao, Yanghua},
  journal={Preprint},
  year={2024}
}

@inproceedings{huang2024emotionalrag,
  title={Emotional RAG: Enhancing role-playing agents through emotional retrieval},
  author={Huang, Le and Lan, Hengzhi and Sun, Zijun and Shi, Chuan and Bai, Ting},
  booktitle={2024 IEEE International Conference on Knowledge Graph (ICKG)},
  pages={120--127},
  year={2024},
  organization={IEEE}
}

@article{chhikara2025mem0,
  title={Mem0: Building production-ready ai agents with scalable long-term memory},
  author={Chhikara, Prateek and Khant, Dev and Aryan, Saket and Singh, Taranjeet and Yadav, Deshraj},
  journal={arXiv preprint arXiv:2504.19413},
  year={2025}
}

@inproceedings{jiang2024personallm,
  title={PersonaLLM: Investigating the ability of large language models to express personality traits},
  author={Jiang, Hang and Zhang, Xiajie and Cao, Xubo and Breazeal, Cynthia and Roy, Deb and Kabbara, Jad},
  booktitle={Findings of the association for computational linguistics: NAACL 2024},
  pages={3605--3627},
  year={2024}
}

@article{serapio2023personality,
  title={Personality traits in large language models},
  author={Serapio-Garc{\'\i}a, Greg and Safdari, Mustafa and Crepy, Cl{\'e}ment and Sun, Luning and Fitz, Stephen and Romero, Peter and Abdulhai, Marwa and Faust, Aleksandra and Matari{\'c}, Maja},
  journal={arXiv preprint arXiv:2307.00184},
  year={2023}
}

@article{zou2023representation,
  title={Representation engineering: A top-down approach to ai transparency},
  author={Zou, Andy and Phan, Long and Chen, Sarah and Campbell, James and Guo, Phillip and Ren, Richard and Pan, Alexander and Yin, Xuwang and Mazeika, Mantas and Dombrowski, Ann-Kathrin and others},
  journal={arXiv preprint arXiv:2310.01405},
  year={2023}
}

@article{chen2025persona,
  title={Persona vectors: Monitoring and controlling character traits in language models},
  author={Chen, Runjin and Arditi, Andy and Sleight, Henry and Evans, Owain and Lindsey, Jack},
  journal={arXiv preprint arXiv:2507.21509},
  year={2025}
}

@inproceedings{han2026personalityillusion,
  title        = {The Personality Illusion: Revealing Dissociation Between Self-Reports and Behavior in LLMs},
  author       = {Han, Pengrui and Kocielnik, Rafal D. and Song, Peiyang and Debnath, Ramit and Mobbs, Dean and Anandkumar, Anima and Alvarez, R. Michael},
  booktitle    = {Proceedings of the International Conference on Machine Learning (ICML)},
  year         = {2026},
  note         = {arXiv:2509.03730}
}

@article{song2025mischaracterize,
  title        = {Human Psychometric Questionnaires Mischaracterize LLM Psychology: Evidence from Generation Behavior},
  author       = {Song, Woojung and Choi, Dongmin and Park, Yoonah and Han, Jongwook and Jo, Yohan},
  journal      = {arXiv preprint arXiv:2509.10078},
  year         = {2025}
}

@inproceedings{wang2026memorydriven,
  author    = {Wang, K. and You, H. and Zhang, Y. and Wang, Z.},
  title     = {Memory-Driven Role-Playing: Evaluation and Enhancement of Persona Knowledge Utilization in {LLMs}},
  booktitle = {Findings of the Association for Computational Linguistics: ACL 2026},
  month     = {July},
  year      = {2026},
  pages     = {23475--23510}
}

@article{tang2026reveriemem,
  author  = {Tang, X. and Zhang, J. and Yang, Z. and Tang, Y. and Li, S. and Lai, L. and Yang, Z.},
  title   = {Staying In Character: Perspective-Bounded Memory For Book-Based Role-Playing Agents},
  journal = {arXiv preprint arXiv:2606.25632},
  year    = {2026},
  month   = {June}
}

@article{mohammad2025nrcvad,
  title   = {{NRC VAD Lexicon v2}: Norms for Valence, Arousal, and Dominance for over 55k English Terms},
  author  = {Mohammad, Saif M.},
  journal = {arXiv preprint arXiv:2503.23547},
  year    = {2025},
  doi     = {10.48550/arXiv.2503.23547},
  url     = {https://arxiv.org/abs/2503.23547}
}

@inproceedings{maharana2024evaluating,
  title     = {Evaluating Very Long-Term Conversational Memory of {LLM} Agents},
  author    = {Maharana, Adyasha and
               Lee, Dong-Ho and
               Tulyakov, Sergey and
               Bansal, Mohit and
               Barbieri, Francesco and
               Fang, Yuwei},
  booktitle = {Proceedings of the 62nd Annual Meeting of the Association for Computational Linguistics (Volume 1: Long Papers)},
  year      = {2024},
  month     = aug,
  pages     = {13851--13870},
  address   = {Bangkok, Thailand},
  publisher = {Association for Computational Linguistics},
  doi       = {10.18653/v1/2024.acl-long.747},
  url       = {https://aclanthology.org/2024.acl-long.747/}
}

@article{goldberg1992development,
  title   = {The Development of Markers for the Big-Five Factor Structure},
  author  = {Goldberg, Lewis R.},
  journal = {Psychological Assessment},
  year    = {1992},
  volume  = {4},
  number  = {1},
  pages   = {26--42},
  doi     = {10.1037/1040-3590.4.1.26}
}

@article{zhang2026beyond,
  title={Beyond Preset Identities: How Agents Form Stances and Boundaries in Generative Societies},
  author={Zhang, Hanzhong and Song, Siyang and Wang, Jindong},
  journal={arXiv preprint arXiv:2603.23406},
  year={2026}
}

@incollection{schapiro1953style,
  author    = {Schapiro, Meyer},
  title     = {Style},
  booktitle = {Anthropology Today: An Encyclopedic Inventory},
  editor    = {Kroeber, A. L.},
  publisher = {University of Chicago Press},
  address   = {Chicago},
  year      = {1953},
  pages     = {287--312}
}

@book{cooper1907theories,
  editor    = {Cooper, Lane},
  title     = {Theories of Style, with Especial Reference to Prose Composition: Essays, Excerpts, and Translations},
  publisher = {The Macmillan Company},
  address   = {New York},
  year      = {1907}
}

@book{possati2022unconscious,
  author    = {Possati, Luca M.},
  title     = {Unconscious Networks: Philosophy, Psychoanalysis, and Artificial Intelligence},
  publisher = {Routledge},
  address   = {New York},
  year      = {2022},
  doi       = {10.4324/9781003345572}
}

@inproceedings{sun2025idiosyncrasies,
  author    = {Sun, Mingjie and Yin, Yida and Xu, Zhiqiu and Kolter, J. Zico and Liu, Zhuang},
  title     = {Idiosyncrasies in Large Language Models},
  booktitle = {Proceedings of the 42nd International Conference on Machine Learning},
  series    = {Proceedings of Machine Learning Research},
  volume    = {267},
  pages     = {57854--57885},
  publisher = {PMLR},
  year      = {2025},
  url       = {https://proceedings.mlr.press/v267/sun25z.html}
}

@article{juzek2026lexical,
  author  = {Juzek, Thomas Stephan},
  title   = {{AI}-Associated Lexical Shifts Across 34 Languages: Cross-Lingual Convergence and Diachronic Uptake in News Writing},
  journal = {arXiv preprint arXiv:2605.25358},
  year    = {2026},
  doi     = {10.48550/arXiv.2605.25358},
  url     = {https://arxiv.org/abs/2605.25358}
}

@incollection{merleauponty1964indirect,
  author    = {Merleau-Ponty, Maurice},
  title     = {Indirect Language and the Voices of Silence},
  booktitle = {Signs},
  translator = {McCleary, Richard C.},
  publisher = {Northwestern University Press},
  address   = {Evanston, IL},
  year      = {1964},
  pages     = {39--83}
}

@book{hayles2005mymother,
  author    = {Hayles, N. Katherine},
  title     = {My Mother Was a Computer: Digital Subjects and Literary Texts},
  publisher = {University of Chicago Press},
  address   = {Chicago},
  year      = {2005}
}

@book{legge1882yiking,
  author    = {Legge, James},
  title     = {The Sacred Books of China: The Texts of Confucianism, Part II: The Y{\^i} King},
  editor    = {M{\"u}ller, F. Max},
  series    = {The Sacred Books of the East},
  volume    = {16},
  publisher = {Clarendon Press},
  address   = {Oxford},
  year      = {1882}
}

@book{sushi1986wenji,
  author    = {{Su Shi}},
  title     = {Su Shi Wen Ji},
  editor    = {{Kong Fanli}},
  publisher = {Zhonghua Shuju},
  address   = {Beijing},
  year      = {1986},
  note      = {In Chinese; includes ``Da Zhang Wenqian Shu''}
}

@article{shumailov2024collapse,
  author  = {Shumailov, Ilia and Shumaylov, Zakhar and Zhao, Yiren and Papernot, Nicolas and Anderson, Ross and Gal, Yarin},
  title   = {{AI} Models Collapse When Trained on Recursively Generated Data},
  journal = {Nature},
  year    = {2024},
  volume  = {631},
  pages   = {755--759},
  doi     = {10.1038/s41586-024-07566-y}
}

@article{inkpin2019style,
  author  = {Inkpin, Andrew},
  title   = {Merleau-Ponty and the Significance of Style},
  journal = {European Journal of Philosophy},
  year    = {2019},
  volume  = {27},
  number  = {2},
  pages   = {468--483},
  doi     = {10.1111/ejop.12430}
}

@book{stanislavski2008actorswork,
  author     = {Stanislavski, Konstantin},
  title      = {An Actor's Work: A Student's Diary},
  editor     = {Benedetti, Jean},
  translator = {Benedetti, Jean},
  publisher  = {Routledge},
  address    = {London and New York},
  year       = {2008}
}

@book{stanislavski2010role,
  author     = {Stanislavski, Konstantin},
  title      = {An Actor's Work on a Role},
  editor     = {Benedetti, Jean},
  translator = {Benedetti, Jean},
  publisher  = {Routledge},
  address    = {London and New York},
  year       = {2010}
}

@book{brecht1964theatre,
  author     = {Brecht, Bertolt},
  title      = {Brecht on Theatre: The Development of an Aesthetic},
  editor     = {Willett, John},
  translator = {Willett, John},
  publisher  = {Hill and Wang},
  address    = {New York},
  year       = {1964}
}

@misc{openpsychometrics2018ipip,
  author       = {{Open-Source Psychometrics Project}},
  title        = {Answers to the {IPIP} Big Five Factor Markers},
  year         = {2018},
  howpublished = {Open psychology data: Raw data from online personality tests},
  note         = {IPIP-FFM-data-8Nov2018; accessed 1 August 2026},
  url          = {https://openpsychometrics.org/_rawdata/}
}

@inproceedings{wang2025coser,
  title={CoSER: Coordinating LLM-Based Persona Simulation of Established Roles.},
  author={Wang, Xintao and Wang, Heng and Zhang, Yifei and Yuan, Xinfeng and Xu, Rui and Huang, Jen-tse and Yuan, Siyu and Guo, Haoran and Chen, Jiangjie and Zhou, Shuchang and others},
  booktitle={ICML},
  year={2025}
}

@inproceedings{liu2025cogdual,
  title={CogDual: Enhancing dual cognition of LLMs via reinforcement learning with implicit rule-based rewards},
  author={Liu, Cheng and Lu, Yifei and Ye, Fanghua and Li, Jian and Chen, Xingyu and Ren, Feiliang and Tu, Zhaopeng and Li, Xiaolong},
  booktitle={Proceedings of the 2025 conference on empirical methods in natural language processing},
  pages={27295--27324},
  year={2025}
}

@inproceedings{du2026her,
  title={Her: Human-like reasoning and reinforcement learning for llm role-playing},
  author={Du, Chengyu and Wang, Xintao and Chen, Aili and Li, Weiyuan and Xu, Rui and Liu, Junteng and Huang, Zishan and Tian, Rong and Sun, Zijun and Li, Yuhao and others},
  booktitle={Findings of the Association for Computational Linguistics: ACL 2026},
  pages={25725--25762},
  year={2026}
}
\bibliographystyle{iclr2027_conference}

\appendix
\section*{Appendix Overview}

The supplementary material contains one conceptual appendix followed by eight technical appendices. Appendix~A presents the broader motivation behind PersMem and the design that we hope to develop in future work. It considers how personality-dependent memory processing may provide a stable experiential basis from which an agent responds across interactions. The discussion is not a literal description of the current PersMem implementation. Instead, it explains the ideas that motivated the present design and outlines a broader direction in which memory, personality, and character expression can be treated as parts of the same agent architecture.

Appendices~B--I provide the technical definitions, implementation details, and experimental settings used in this paper:
\begin{itemize}
    \item Appendix~A: Toward a Characteristic Voice: A Conceptual and Philosophical Perspective on Personality-Grounded Responses (p. 1);
    \item Appendix~B: Fixed Hyperparameters (p. 3);
    \item Appendix~C: Attachment Instantiation and Method Details (p. 4);
    \item Appendix~D: Big Five Instantiation (p. 7);
    \item Appendix~E: Experiment~1 Details (p. 9);
    \item Appendix~F: Experiment~2 Details (p. 13);
    \item Appendix~G: LoCoMo Factual QA Details (p. 15);
    \item Appendix~H: Adversarial Positive Framing Details (p. 16);
    \item Appendix~I: CoSER Role-Playing Evaluation Details (p.~\pageref{app:coser}).
\end{itemize}

Readers interested mainly in the implementation and reproducibility of PersMem may begin with Appendix~B.

\FloatBarrier
\section{Appendix A. Toward a Characteristic Voice: A Conceptual and Philosophical Perspective on Personality-Grounded Responses}

``Long experience with art has established as a plausible principle the notion that an individual style is a personal expression'' \cite{schapiro1953style}. Throughout the history and practice of literary criticism, we have become accustomed to associating a person's written works with that person's personality. Confucius, a Chinese thinker who lived at roughly the same time as Socrates, had already observed that ``The virtuous are sparing in speech; the rash are verbose; the calumnious are evasive; the unprincipled are feeble in speech'' (\textit{The Book of Changes}, ``Xi Ci II''; translation adapted by the authors) \cite{legge1882yiking}. By the eleventh century, the renowned writer Su Shi summarised this idea as ``One's writing reflects one's character'' \cite{sushi1986wenji}. In Europe, Buffon's address upon his admission to the Académie Française, ``the style is the man himself'' \cite{cooper1907theories}, continues to influence generations of literary creators and researchers.

But what about artificial intelligence?

Artificial intelligence appears to have a style. Across online communities, we can easily find large numbers of comments from users of conversational agents about their “language styles,” most of them negative. These agents seem to prefer certain fixed sentence patterns, combinations, and modes of expression, stubbornly and confidently maintaining their own ``characteristics.'' They also thoughtfully offer users a choice: you can make ``my'' expression warmer or colder. Every change in output style provokes intense discussion among communities and users. Within a remarkably short period, the ``styles'' of these agents have not only taken shape, but have even acquired their own admirers and critics. Yet are these habitual ways of responding sufficient to constitute a genuine ``style''?

In some respects, agents may be closer to ``style'' than we imagine. One point of agreement among contemporary literary critics is that style is closely related to the collective creative tendencies, shared sources, modes of training, and aesthetic preferences of the period in which a creator lives. Even the greatest genius develops a style through communication and interaction within the surrounding world, and often within a group of closely connected individuals and communities. Artificial intelligence has performed remarkably well along this path. First, AI takes the vast vocabulary of the contemporary Internet as its language teacher. Through repeated training on these materials and their reproduction as conversational text, the historical characteristics embedded within them become the preconscious background of the agent's language. The desires suppressed or exposed within these texts may even become part of the unconscious of artificial intelligence \cite{possati2022unconscious}. 

Second, mutual distillation among major AI companies has already organised these ``giant'' agents into a substantial stylistic community. While relying on ever-increasing scale to refresh one benchmark after another, agents absorb one another's habits of textual expression. The significant differences in ``linguistic habits'' among large language models previously identified by researchers \cite{sun2025idiosyncrasies} are now being overshadowed by the prospect of model collapse. At the same time, as AI-generated content increasingly enters the data ecosystem used to train later models, repeated training on generated data may reduce distributional diversity and contribute to model collapse \cite{shumailov2024collapse}. In this sense, convergence may affect not only what agents say, but also how they say it. An ``artificial hive mind'' not only governs the ``content'' itself, but also erodes the ways in which that content is expressed. Finally, large language model agents are not only transforming one another's habits of expression, but also those of their interlocutors---the users of these models \cite{juzek2026lexical}. Some users discover through repeated interaction that certain forms of expression enable their models to ``understand'' them more fully and guide the models toward results that they find more acceptable. Other users may regard chatbots as more organised, logical, and intelligent examples, and therefore consciously adopt chatbot-like language, or make only minor revisions to text generated by agents. The interlocutors of an agent could have provided a rich and diverse source of language, supplying the agent with fresh references that differ from standardised expression. Yet as the expressive habits of these interlocutors become increasingly AI-like, the agent's own style can only become more fixed. From these three perspectives, artificial intelligence appears to be moving in the right direction in its work toward ``style.''

Yet all of these forms of style may depend on one premise that artificial intelligence has not shown us: a creator. This may sound anthropocentric, but that is precisely where the problem lies. It is difficult to regard a series of random combinations, or a Brownian motion of words, as a coherent creative activity. Nor would we describe the Great Barrier Reef and Yellowstone National Park as possessing different ``styles''---the textual responses of artificial intelligence resemble such an artificial landscape. Under the constraints imposed by descriptive instructions, an agent attempts to maintain an image, which it understands as a kind of style---and perhaps this is also true of most human writers. However, artificial intelligence cannot guarantee its own stability. In extended interactions, an agent may lose persona or stylistic consistency \cite{kovac2024stick,abdulhai2025consistently,ding2026contextecho}, a failure sometimes described informally in online communities as ``goblin syndrome.''\footnote{We use ``goblin syndrome'' only as an informal community expression for conspicuous degeneration or incoherence in long-form model output; it is not an established technical term.} At a more fundamental level, there is nothing within artificial intelligence that can confirm that the texts it creates remain ``stable.'' Furthermore, artificial intelligence has no perception of the image that it is producing. Human creators often experience their own writing while creating it. Beyond recognising it as a particular style, they are affected by the expressive power of that style, which further stimulates their sympathetic imagination and encourages continued writing in that style---or causes the style to collapse. Whether the style is reinforced or dissolved, we can often trust the process. This is the ``stabilising'' function that personality performs for style.

Researchers who support the view that artificial intelligence possesses style often cite Merleau-Ponty's account of style as a ``system of equivalences'' (\textit{système d'équivalence}) or a ``coherent deformation'' (\textit{déformation cohérente}) \cite{merleauponty1964indirect}. They further argue that AI-generated text already satisfies both requirements: the creative subject has established an equivalence between expression and signs and has also achieved the continuous generation of a system of signs \cite{merleauponty1964indirect,inkpin2019style}. Yet although Merleau-Ponty appears to remove the individual author, he reaffirms authorship within ``style.'' Style is not the final purpose of a text, but a mediating layer that leads readers toward an understanding of themselves, of others, and of the world. What these relationships reveal is precisely the author's emotional patterns and perceptual structures. Although personality governs style, the reader is no longer merely an admirer kneeling before the personality of a genius. By recognising a style and having experiences evoked through it, the reader instead becomes the author's interlocutor. In other words, within the open world of a stylistic text, the reader also becomes a co-author of ``the text'' as it exists within the reader's own experience of reading. This is the ``mediating'' function that personality performs for style. Artificial intelligence can only extract rhetorical tendencies from large quantities of data, and the resulting texts are therefore products of style, whereas ``style'' itself ultimately signifies the expression of personality. As N. Katherine Hayles writes, ``Language, emerging from the operations of the unconscious figured as a Turing machine, creates expressions of desire that in their origin are always already interpenetrated by the mechanistic, no matter how human they seem'' \cite{hayles2005mymother}.

From this perspective, at least until the emergence of strong artificial intelligence, AI-generated texts may therefore remain unable to possess style as a distinctly human form of wealth. Yet restricting artificial intelligence to human forms of style would itself amount to ``being stuck in one's ways.'' It is difficult to manufacture something that we do not yet understand: what personality consists of and how it is formed remain unresolved questions. However, we do not need to manufacture a genuine personality. What matters more is that a synthetic personality should at least carry the basic functions that personality performs in relation to style: it should support the stability and interpretability of habitual expression as a ``stabiliser,'' while also governing expression in a way that enables dialogue and communication with others as a ``mediator.'' Grounded in technological pragmatism, PersMem adopts a synthetic approach and attempts to make progress toward these two functions.

PersMem does not construct personhood. Instead, it is closer to constructing a character. In role-playing games, we commonly assign characters different settings and backgrounds, which are also reflected in their attributes, abilities, and tendencies. By initially specifying a quantitatively structured ``personality scale,'' PersMem attempts to perform such a personality-based character. However, PersMem does not begin this performance through rhetoric. It begins through memory.

PersMem resembles a skilled actor working within Stanislavski's system. Before producing a line that is ``consistent with the character,'' it must complete three necessary and principled preparations. First, it must show absolute respect for the ``playwright''---in this context, the requirements established by its interlocutor for the conversation. Within Stanislavski's theatre, the playwright always occupies the central position. This means that before the agent fills the character through its own ``imagination,'' it must treat the material provided by the playwright---the remembered texts of previous dialogue---as its primary principle. The character may extend beyond these materials, but these materials must always occupy the central and decisive position within the character. The actor's individual creation and interpretation can be established only after being tested against the playwright and the dramatic material. Second, the actor must write a ``character biography.'' It must actively search through all available materials---that is, the memories of its conversations---for every trace of the character: preferences, habits, life experiences, emotional experiences, and unforgettable events that have shaped its ``personality'' \cite{stanislavski2008actorswork,stanislavski2010role}. PersMem first searches the repository of materials and organises these texts as ``experience.'' It then constructs episodes and fragments from the character's life and, after receiving the playwright's approval, reinforces these biographical fragments as the character's defining characteristics. On the theatrical stage, these characteristics may take the form of a distinctive voice, particular bodily habits, recurring facial expressions, or even clothing. In an agent's dialogue, they first appear as its ``characteristic voice'' \footnote{We use ``characteristic voice'' as a design concept for responses whose continuity is grounded in personality-dependent experience and memory processing, rather than merely in recurring surface expressions. It is not intended as an established psycholinguistic construct.}. Third, the actor must reawaken ``experience'' and ``emotion'' from within its own life. An actor may never have experienced the separation of Romeo and Juliet, yet may still possess other experiences of parting. The actor may never have fallen in love, but liking, jealousy, longing, desire, and sadness remain present throughout human life. For the actor, what matters is the ability to awaken and combine these emotions and sensations while performing a role. In the agent's performance, this corresponds to a selective and weighted secondary synthesis of its memories. Unlike the ``character biography,'' which is organised according to the playwright's requirements, this process is governed by the agent's pre-defined personality scale. The agent's personality-based character does not merely represent the character that it performs; it also indicates that this particular agent-actor is especially suited to performing that character. This means that the agent-actor already possesses the strongest tendencies in the dimensions required by the role. It will therefore tend to select memories that are more consistent with its own personality traits, and thus with the personality traits of the character, while neglecting aspects in which it is less suited to the role. Othello is certainly a noble man, but he is also a husband overwhelmed by passion and jealousy. He may be warm-hearted, but this is not important to the role. The actor therefore does not need to awaken the warm-hearted aspects of the self, and the secondary synthesis of memory will not devote much attention to memories related to kindness or generosity. Othello is certainly not a rational man. The actor will therefore resist wise memories and rational reflections. Even if the actor is cautious and rational in ordinary life, the performance requires these states of mind to remain dormant in order to avoid a distancing effect  \cite{brecht1964theatre}. PersMem, in effect, allows the agent most like Othello to perform Othello. Before beginning the performance, the agent is already almost an Othello---at least according to its personality scale.

We therefore obtain our ``character.'' First, the character follows the user's initial settings, requirements, and conversational texts. Second, the performer of the character---the agent---constructs, under the user's guidance, a background narrative combining major life events with fragments of everyday life, thereby reinforcing the character's mode of expression. Finally, this performer is not a completely neutral, empty, and universally capable actor. It is the actor most similar to the character. From the beginning, the performer possesses particular talents, abilities, and tendencies for performing this role. It draws upon these tendencies to mobilise the memories that it can most readily activate and that are also most consistent with the role, transforming them into the central material through which the character is expressed. The agent first performs an actor who resembles a particular character, and then performs the character that this actor is meant to play. Through this interaction, the stability of expression no longer depends on accidental similarities among rhetorical forms, but on a strongly organised and subjectively grounded memory. At the same time, the resulting expression becomes a mediating platform through which the user can communicate with the remembered subject performed by the agent. In this sense, PersMem reveals one possible path toward dialogue with a characteristic voice.

\FloatBarrier
\section{Appendix B. Fixed Hyperparameters}
\label{app:hparams}

Table~\ref{tab:fixed-hparams} lists the fixed coefficients used by PersMem. These values remain unchanged during calibration and evaluation. The attachment and Big Five instantiations share the same base coefficients wherever the corresponding operation is used, but differ in their theory-specific mappings. Appendices~C and~D specify the calibrated gains and mappings for the two instantiations.

\begin{table}[t]
\centering
\caption{Fixed coefficients used by PersMem. Calibrated retrieval gains and the Big Five sampling temperature are defined separately in Appendices~C and~D.}
\label{tab:fixed-hparams}
\footnotesize
\setlength{\tabcolsep}{3pt}
\begin{tabular}{@{}lp{0.56\columnwidth}r@{}}
\toprule
Symbol & Role & Value \\
\midrule
$\kappa_v$ & valence adjustment & $0.30$ \\
$\kappa_a$ & negative-event arousal increase & $0.20$ \\
$\kappa_{\mathrm{damp}}$ & avoidance-related arousal damping & $0.15$ \\
$\kappa_p$ & dominance increase & $0.10$ \\
$\kappa_q$ & dominance decrease & $0.10$ \\
$r_0$ & base memory decay rate & $10^{-4}\,\mathrm{s}^{-1}$ \\
$c$ & arousal-dependent decay coefficient & $0.85$ \\
$\rho$ & negative-memory retention coefficient & $0.40$ \\
$\epsilon_{\mathrm{ret}}$ & minimum decay multiplier & $10^{-3}$ \\
$\alpha$ & semantic score weight & $0.50$ \\
$\beta$ & base affective score weight & $0.30$ \\
$\gamma$ & retention score weight & $0.20$ \\
$\xi$ & trait-dependent affective scaling & $0.30$ \\
$w_a$ & arousal-match weight & $0.30$ \\
$w_v$ & valence-match weight & $0.50$ \\
$w_d$ & dominance-match weight & $0.20$ \\
$b_{\mathrm{fa}}$ & functional-avoidance offset & $0.50$ \\
$\lambda_I$ & redirection-gate slope & $6.0$ \\
$\theta_0$ & base gate threshold & $0.50$ \\
$\eta$ & trait-dependent threshold reduction & $0.20$ \\
$\nu$ & arousal contribution to the gate & $1.0$ \\
$\lambda$ & affective-state update rate & $0.40$ \\
$\tau_{\mathrm{att}}$ & attachment sampling temperature & $1.0$ \\
$T$ & default active-retrieval step budget & $4$ \\
\bottomrule
\end{tabular}
\end{table}

\FloatBarrier
\section{Appendix C. Attachment Instantiation and Method Details}
\label{app:attachment-method}

This appendix specifies the attachment instantiation of PersMem. Attachment theory represents personality using two continuous dimensions, attachment anxiety and attachment avoidance \cite{bartholomew1991attachment,brennan1998selfreport}. PersMem maps this fixed two-dimensional personality vector to parameters controlling affective appraisal, memory retention, passive affect-driven memory retrieval, and personality-conditioned redirection within active goal-driven memory retrieval. Neither the active retrieval controller nor the response model receives the personality vector.

\subsection{Personality and Memory Representation}

The attachment instantiation represents personality as
$\pi=(\pi_{\mathrm{anx}},\pi_{\mathrm{avo}})\in[0,1]^2$, where
$\pi_{\mathrm{anx}}$ and $\pi_{\mathrm{avo}}$ denote attachment anxiety and avoidance, respectively. The four evaluated attachment profiles are
\begin{equation}
\begin{aligned}
\pi_{\mathrm{secure}}
&=(0.10,0.10),&
\pi_{\mathrm{anxious}}
&=(0.85,0.15),\\
\pi_{\mathrm{avoidant}}
&=(0.15,0.85),&
\pi_{\mathrm{fearful}}
&=(0.80,0.80).
\end{aligned}
\label{eq:attachment-profiles}
\end{equation}
These values are representative operating points near the corners of the continuous anxiety--avoidance space rather than estimates of individual users.

Affective information is represented by a valence-arousal-dominance vector $e=(v,a,d)$, where $v\in[-1,1]$ denotes valence, $a\in[0,1]$ denotes arousal, and $d\in[0,1]$ denotes dominance. A memory item is represented as
$m=(x_m,\mathbf z_m,e_m,t_m,\mathrm{spec}_m)$, where $x_m$ is the memory text, $\mathbf z_m$ is its normalised sentence embedding, $e_m=(v_m,a_m,d_m)$ is its affective annotation, $t_m$ is its timestamp, and $\mathrm{spec}_m$ indicates whether the memory is specific or overgeneral. We use \texttt{paraphrase-multilingual-MiniLM-L12-v2} to obtain the sentence embeddings.

The predefined specificity labels are used during calibration and by the functional-avoidance term. The cross-label evaluation instead computes its reported outcomes using a separately generated blind label set. Newly stored conversational exchanges receive the default label \texttt{specific}. The functional-avoidance term therefore applies only to memories for which a specificity label is available.

We use $[x]_+=\max(x,0)$ to denote the positive part of $x$, $\sigma$ to denote the Sigmoid function, and $\operatorname{clip}_{[l,u]}$ to denote clipping to the interval $[l,u]$.

\subsection{Affective Appraisal and Current Affective State}

A low-temperature LLM annotator maps each text input to a raw VAD vector
$\tilde e=(\tilde v,\tilde a,\tilde d)$. The annotator receives at most 500 characters, uses a temperature of $0.1$ and a maximum output length of 80 tokens, and returns a JSON object. Invalid or missing outputs are replaced with $(0,0,0.5)$, after which each component is clipped to its valid range. The model used by the annotator is specified in the corresponding experimental setup.

The attachment-dependent appraisal mapping $\mathcal A_\pi$ converts the raw VAD vector $\tilde e$ into the adjusted vector $e'=(v',a',d')$. The coefficients $\kappa_v$, $\kappa_a$, $\kappa_{\mathrm{damp}}$, $\kappa_p$, and $\kappa_q$ control the personality-dependent adjustments to valence, arousal, and dominance:
\begin{equation}
\begin{aligned}
v'
&=
\operatorname{clip}_{[-1,1]}\Bigl(
\tilde v
+\kappa_v(1-\pi_{\mathrm{anx}})[\tilde v]_+\\
&\qquad
-\kappa_v\pi_{\mathrm{anx}}[-\tilde v]_+
\Bigr),\\
a'
&=
\operatorname{clip}_{[0,1]}\Bigl(
\tilde a
+\kappa_a\pi_{\mathrm{anx}}[-\tilde v]_+\\
&\qquad
-\kappa_{\mathrm{damp}}\pi_{\mathrm{avo}}\tilde a
\Bigr),\\
d'
&=
\operatorname{clip}_{[0,1]}\Bigl(
\tilde d
+\kappa_p\pi_{\mathrm{avo}}(1-\tilde d)\\
&\qquad
-\kappa_q\pi_{\mathrm{anx}}\tilde d
\Bigr).
\end{aligned}
\label{eq:attachment-appraisal}
\end{equation}
Higher attachment anxiety makes negative valence more negative, increases arousal for negative events, and reduces dominance. Higher attachment avoidance reduces arousal and increases dominance.

PersMem maintains a dynamic VAD vector representing the agent's current affective state. The initial state is $s_0=(0,0,0.5)$. At interaction turn $t$, $s_t^{\mathrm{in}}$ denotes the affective state after appraising the current user input, while $s_t^{\mathrm{out}}$ denotes the state after appraising the generated reply. The parameter $\lambda\in[0,1]$ controls the contribution of the newly appraised text. The adjusted VAD vectors of the input and reply are denoted by $e_t^{\mathrm{in}}$ and $e_t^{\mathrm{out}}$, respectively, and their mean $e_{m,t}$ is assigned to the completed exchange:
\begin{equation}
\begin{aligned}
s_t^{\mathrm{in}}
&=
(1-\lambda)s_{t-1}^{\mathrm{out}}
+\lambda e_t^{\mathrm{in}},\\
s_t^{\mathrm{out}}
&=
(1-\lambda)s_t^{\mathrm{in}}
+\lambda e_t^{\mathrm{out}},\\
e_t^{\mathrm{in}}
&=
\mathcal A_\pi(\tilde e_t^{\mathrm{in}}),\qquad
e_t^{\mathrm{out}}
=
\mathcal A_\pi(\tilde e_t^{\mathrm{out}}),\\
e_{m,t}
&=
\frac{e_t^{\mathrm{in}}+e_t^{\mathrm{out}}}{2}.
\end{aligned}
\label{eq:attachment-state}
\end{equation}
Memory retrieval at turn $t$ uses the state $s_t^{\mathrm{in}}$. Each component of the updated state is rounded to three decimal places.

\subsection{Memory Retention}

For each stored memory $m$, PersMem computes a retention value representing how accessible the memory remains at time $t$. The base decay rate is denoted by $r_0$, $c$ controls the effect of memory arousal on decay, $\rho$ controls the anxiety-dependent slowing of decay for negative memories, and $\epsilon_{\mathrm{ret}}$ provides a lower bound for the personality-dependent decay multiplier:
\begin{equation}
\begin{aligned}
r(m;\pi)
&=
r_0(1-ca_m)
\max\Bigl(
\epsilon_{\mathrm{ret}},
1-\rho\pi_{\mathrm{anx}}[-v_m]_+
\Bigr),\\
\operatorname{ret}(m,t;\pi)
&=
\exp\bigl(-(t-t_m)r(m;\pi)\bigr).
\end{aligned}
\label{eq:attachment-retention}
\end{equation}
Here, $r(m;\pi)$ is the personality-dependent decay rate of memory $m$, and $\operatorname{ret}(m,t;\pi)$ is its retention value at time $t$. Because $c\in[0,1)$ and $a_m\in[0,1]$, the arousal-dependent multiplier remains positive. The personality-dependent multiplier is lower-bounded by $\epsilon_{\mathrm{ret}}=10^{-3}$. Over the declared personality and valence ranges, its unclipped value is at least $0.60$, so this lower bound is inactive in the reported experiments. Higher memory arousal slows forgetting, while higher attachment anxiety further slows the forgetting of negative memories.

\subsection{Passive Affect-Driven Memory Retrieval}

Given the current search query $q$ and affective state $s=(v_s,a_s,d_s)$, passive affect-driven memory retrieval assigns a score to each available memory $m$. The score combines the semantic similarity $\operatorname{sem}(q,m)$ between the query and the memory, the retention value $\operatorname{ret}(m,t;\pi)$, and the attachment-dependent affective relation score $R(s,m;\pi)$. The coefficients $\alpha$, $\beta$, and $\gamma$ control the contributions of semantic similarity, affective relation, and retention, respectively, while $\xi$ controls the effect of attachment anxiety on the affective-relation weight:
\begin{equation}
\begin{aligned}
S_{\mathrm{pas}}(m)
&=
\alpha\,\operatorname{sem}(q,m)
+\beta_\pi R(s,m;\pi)\\
&\qquad
+\gamma\,\operatorname{ret}(m,t;\pi),\\
\operatorname{sem}(q,m)
&=
\mathbf z_q^\top\mathbf z_m,\qquad
\beta_\pi
=
\beta(1+\xi\pi_{\mathrm{anx}}),
\end{aligned}
\label{eq:attachment-passive-score}
\end{equation}
where $\mathbf z_q$ is the normalised sentence embedding of the current query and $S_{\mathrm{pas}}(m)$ is the resulting passive-retrieval score.

The affective relation score uses the weights $w_a$, $w_v$, and $w_d$ for arousal, valence, and dominance consistency, respectively. The gains $\mu$ and $\mu_d$ control the additional effects of attachment anxiety on negative memories and attachment avoidance on high-dominance memories. The gain $\kappa_{\mathrm{fa}}$ controls the functional-avoidance term, while $b_{\mathrm{fa}}$ is its base value:
\begin{equation}
\begin{aligned}
R(s,m;\pi)
={}&
w_a(1-|a_s-a_m|)
+w_v\frac{1+v_sv_m}{2}\\
&+
w_d(1-|d_s-d_m|)
+\mu\pi_{\mathrm{anx}}[-v_m]_+\\
&+
\mu_d\pi_{\mathrm{avo}}d_m
+\mathbb I_m^{\mathrm{og}}
\Delta_{\mathrm{fa}}(s,\pi),\\
\Delta_{\mathrm{fa}}(s,\pi)
={}&
\kappa_{\mathrm{fa}}
\left(
\pi_{\mathrm{avo}}
+\frac{1}{2}\pi_{\mathrm{anx}}
\right)\\
&\times
\left(
b_{\mathrm{fa}}
+[-v_s]_+a_s
\right).
\end{aligned}
\label{eq:attachment-affective-score}
\end{equation}
Here, $\mathbb I_m^{\mathrm{og}}$ is $1$ when memory $m$ is labelled as overgeneral and $0$ otherwise. The final term implements the functional-avoidance account of overgeneral autobiographical memory \cite{williams2007autobiographical} and applies only to memories with predefined specificity labels.

Given a candidate set $\mathcal C$ and sampling temperature $\tau>0$, PersMem converts the passive-retrieval scores into the sampling distribution
\begin{equation}
P_{\mathrm{pas}}(m\mid\mathcal C)
=
\frac{
\exp\bigl(S_{\mathrm{pas}}(m)/\tau\bigr)
}{
\sum_{m'\in\mathcal C}
\exp\bigl(S_{\mathrm{pas}}(m')/\tau\bigr)
},
\qquad
m\in\mathcal C,
\label{eq:passive-sampling}
\end{equation}
where $m'$ indexes the memories in $\mathcal C$. This sampling rule is shared by the attachment and Big Five instantiations. The construction of $\mathcal C$ depends on the calling procedure:
\begin{itemize}
\item Conversational passive affect-driven memory retrieval uses the $K$ highest-scoring memories, where $K=\max(2n,4)$ and $n=3$ is the default number of sampled memories.
\item At each active retrieval step, the candidate set contains memories not yet selected during the current active goal-driven memory retrieval.
\item Attachment calibration computes its retrieval statistics over the complete calibration memory store.
\end{itemize}
The attachment instantiation uses the fixed sampling temperature $\tau=\tau_{\mathrm{att}}=1.0$.

\subsection{Active Goal-Driven Memory Retrieval}

When an explicit recall goal $g$ is available, passive affect-driven memory retrieval first produces the initial passive memory set $R^{\mathrm{pas}}$. Active goal-driven memory retrieval then starts with the empty active memory set $R_0^{\mathrm{act}}=\varnothing$ and uses the recall goal as its initial query, $q_0=g$. The passive and active memory sets remain separate during retrieval and are merged after the active process terminates.

Let $\mathcal M$ denote the complete memory store. At active retrieval step $t$, the passive pathway samples one candidate memory $m_t^{\mathrm{pas}}$ from
$\mathcal C_t=\mathcal M\setminus R_t^{\mathrm{act}}$. The gate-firing probability is denoted by $I_t$, and the resulting Bernoulli decision is denoted by $B_t$. The coefficient $\lambda_I$ controls the effect of the difference between the candidate score and the firing threshold, $\nu$ controls the effect of the current arousal $a_t$, $\theta_0$ is the base firing threshold, and $\eta$ controls the effect of attachment anxiety on this threshold:
\begin{equation}
\begin{aligned}
I_t
&=
\sigma\Bigl(
\lambda_I
\bigl[
S_{\mathrm{pas}}(m_t^{\mathrm{pas}})
-\theta_\pi
\bigr]
+\nu a_t
\Bigr),\\
\theta_\pi
&=
\theta_0-\eta\pi_{\mathrm{anx}},\qquad
B_t\sim\operatorname{Bernoulli}(I_t).
\end{aligned}
\label{eq:attachment-redirection-gate}
\end{equation}
The score $S_{\mathrm{pas}}(m_t^{\mathrm{pas}})$ is evaluated before softmax normalisation. Higher attachment anxiety lowers the personality-conditioned threshold $\theta_\pi$ and therefore increases the probability that the gate fires.

When $B_t=1$, PersMem adds $m_t^{\mathrm{pas}}$ to the active memory set, skips semantic retrieval and LLM-based query refinement at the current step, and uses the first 60 characters of the candidate memory as the next query $q_{t+1}$. Active retrieval then continues from this redirected query.

When $B_t=0$, PersMem retrieves the unused memory with the highest semantic similarity to $q_t$ and adds it to the active memory set. An LLM controller then receives the original recall goal, the current search query, and the first 50 characters of each memory accumulated during active goal-driven memory retrieval. The controller receives neither the personality vector $\pi$ nor the current affective state. It uses a temperature of $0.4$ and a maximum output length of 64 tokens, and returns either \texttt{DONE} or a revised search query. When a revised query is returned, PersMem retains the first 80 characters of its first line as $q_{t+1}$.

Active retrieval stops when the controller returns \texttt{DONE} or an empty output, when no unused memory remains, or when the maximum retrieval-step budget $T$ is reached. PersMem then merges and deduplicates the passive and active memory sets:
\begin{equation}
R
=
\operatorname{Deduplicate}
\left(
R^{\mathrm{pas}}\cup R_T^{\mathrm{act}}
\right).
\label{eq:attachment-final-memory-set}
\end{equation}
When no separate controller model is supplied, PersMem uses the same auxiliary model as the affective annotator.

\paragraph{Gate-control modes.}
The ablation includes three gate-control modes. In \emph{personality} mode, PersMem uses the personality-conditioned threshold
$\theta_\pi=\theta_0-\eta\pi_{\mathrm{anx}}$. In \emph{neutral} mode, PersMem retains the same stochastic gate and active retrieval procedure but removes the personality-dependent adjustment to the threshold. In \emph{off} mode, $I_t=0$, so a passively proposed candidate cannot redirect the active retrieval step. All other enabled components remain unchanged across the three modes.

\subsection{Gate Curve}

For the fixed query and memory store used in the gate-curve analysis, we vary only the controlled affective state
$s(a)=(-0.3,a,0.4)$. Let $m^*(a)$ denote the memory with the highest passive-retrieval score under arousal value $a$, and let $H(a)$ denote its corresponding gate-firing probability:
\begin{equation}
\begin{aligned}
m^*(a)
&=
\arg\max_m
S_{\mathrm{pas}}(m\mid s(a)),\\
H(a)
&=
\sigma\Bigl(
\lambda_I
\bigl[
S_{\mathrm{pas}}(m^*(a))
-\theta_\pi
\bigr]
+\nu a
\Bigr).
\end{aligned}
\label{eq:attachment-gate-curve}
\end{equation}
Thus, $H(a)$ represents the gate-firing probability under a controlled arousal sweep rather than the observed redirection frequency over a complete active retrieval process.

\subsection{Controlled Component Switches}

The implementation provides independent switches for personality-dependent affective appraisal (A), memory retention (R), passive affect-driven memory retrieval (P), and personality-conditioned redirection within active retrieval (G), producing a $2^4$ factorial design. A fifth switch for the functional-avoidance specificity term (S) is evaluated separately on memory sets with specificity annotations.

When A is disabled, PersMem retains the profile-independent raw VAD annotation instead of applying the adjustment in Equation~\ref{eq:attachment-appraisal}. When R is disabled, PersMem removes the personality-dependent negative-memory factor while retaining the base time-dependent and arousal-dependent decay terms. When P is disabled, PersMem removes the personality-dependent terms from passive retrieval while retaining semantic similarity, affective matching, memory retention, candidate-set construction, and sampling. When G is disabled, $I_t$ is fixed to zero, so each active retrieval step uses semantic retrieval. When S is disabled, PersMem removes the term containing $\mathbb I_m^{\mathrm{og}}$ while leaving the other passive-retrieval terms unchanged.

A1R1P1G1 denotes the complete personality-dependent system. A value of zero indicates that the corresponding personality-dependent component is disabled. The specificity switch is reported separately because it applies only to memories labelled as specific or overgeneral.

\subsection{Trait-to-Gain Calibration}
\label{sec:attachment-gain-calibration}

The attachment trait-to-gain mapping $g_\phi$, parameterised by $\phi$, is a multilayer perceptron with one hidden layer of eight units, a hyperbolic-tangent activation, and two affine transformations:
\begin{equation}
\begin{aligned}
g_\phi:
(\pi_{\mathrm{anx}},\pi_{\mathrm{avo}})
&\mapsto
(\mu,\mu_d,\kappa_{\mathrm{fa}}),\\
0<\mu,\mu_d&<1,\qquad
0<\kappa_{\mathrm{fa}}<1.2.
\end{aligned}
\label{eq:attachment-gain-map}
\end{equation}
Sigmoid-based output transformations enforce these ranges. The output gains $\mu$, $\mu_d$, and $\kappa_{\mathrm{fa}}$ control the anxiety-dependent negative-memory term, the avoidance-dependent dominance term, and the functional-avoidance term in Equation~\ref{eq:attachment-affective-score}, respectively.

Personality-consistent bias (PCB) measures the mean negativity of the three memories with the highest passive-retrieval scores across the calibration queries. The overgeneral-memory rate (OGM) measures the proportion of sampled memories labelled as overgeneral. Let $\mathcal Q$ denote the calibration-query set, $L$ denote the number of sampling trials per query, and $m_{q,\ell}$ denote the memory sampled for query $q$ in trial $\ell$:
\begin{equation}
\begin{aligned}
\operatorname{PCB}
&=
\frac{1}{|\mathcal Q|}
\sum_{q\in\mathcal Q}
\frac{1}{3}
\sum_{m\in\operatorname{Top3}_{S_{\mathrm{pas}}}(q)}
(-v_m),\\
\operatorname{OGM}
&=
\frac{1}{|\mathcal Q|L}
\sum_{q\in\mathcal Q}
\sum_{\ell=1}^{L}
\mathbb I\bigl[
\mathrm{spec}(m_{q,\ell})
=
\mathrm{overgeneral}
\bigr].
\end{aligned}
\label{eq:attachment-calibration-statistics}
\end{equation}
Here, $\operatorname{Top3}_{S_{\mathrm{pas}}}(q)$ denotes the three memories with the highest passive-retrieval scores for query $q$, and $\mathbb I[\cdot]$ is the indicator function. PCB uses the fixed calibration state
$s_{\mathrm{cal}}=(0,0.30,0.50)$, and attachment calibration uses $L=20$ sampling trials for each query.

Let $b_{\mathrm{PCB}}$ and $b_{\mathrm{OGM}}$ denote the PCB and OGM values obtained after disabling the personality-dependent terms. The prespecified calibration targets are
\begin{equation}
\begin{aligned}
\operatorname{PCB}^*
&=
b_{\mathrm{PCB}}
-0.06
+0.45\pi_{\mathrm{anx}}
+0.04\pi_{\mathrm{avo}},\\
\operatorname{OGM}^*
&=
b_{\mathrm{OGM}}
+0.20\pi_{\mathrm{avo}}
+0.10\pi_{\mathrm{anx}}.
\end{aligned}
\label{eq:attachment-calibration-targets}
\end{equation}
These target relations define the intended directions and scales used for calibration rather than empirical population-level effects.

For numerical feasibility, each target is clipped with a margin of $0.01$ to the range obtained under the direct reference gain settings $(0,0,0)$ and $(1,1,1.2)$. These reference settings directly assign the three gains and therefore bypass the output transformations of $g_\phi$.

Calibration uses the four attachment profiles in Equation~\ref{eq:attachment-profiles} together with the interior personality settings
$(0.50,0.50)$, $(0.30,0.70)$, $(0.70,0.30)$, and $(0.95,0.50)$. Let $\mathcal C_{\mathrm{att}}$ denote this set of eight calibration settings. The calibration objective is
\begin{equation}
\begin{aligned}
\mathcal L_{\mathrm{att}}
={}&
\sum_{j\in\mathcal C_{\mathrm{att}}}
\Bigl[
(\operatorname{PCB}_j-\operatorname{PCB}_j^*)^2\\
&\qquad+
(\operatorname{OGM}_j-\operatorname{OGM}_j^*)^2
\Bigr]
+2\mathcal L_{\mathrm{dom}},\\
\mathcal L_{\mathrm{dom}}
={}&
\left[
\bar d_{\mathrm{secure}}
+0.03
-\bar d_{\mathrm{avoidant}}
\right]_+^2\\
&+
\left[
\bar d_{\mathrm{secure}}
+0.03
-\bar d_{\mathrm{fearful}}
\right]_+^2,
\end{aligned}
\label{eq:attachment-calibration-objective}
\end{equation}
where $\operatorname{PCB}_j$ and $\operatorname{OGM}_j$ denote the statistics obtained for calibration setting $j$, and $\bar d_{\mathrm{secure}}$, $\bar d_{\mathrm{avoidant}}$, and $\bar d_{\mathrm{fearful}}$ denote the mean dominance of the memories retrieved for the corresponding attachment profiles. The term $\mathcal L_{\mathrm{dom}}$ encourages the avoidant and fearful profiles to retrieve memories with higher dominance than the secure profile by a margin of $0.03$.

We fit $g_\phi$ using a Gaussian evolution strategy with 25 generations, a population size of 12, five elites per generation, an initial parameter scale of $0.5$, and a minimum coordinate-wise standard deviation of $0.01$. Random seed 0 is reused across candidate evaluations so that different parameter vectors are evaluated using the same random sequence. After optimisation, the parameter vector with the lowest observed objective value is restored.

The trait-to-gain mapping is fitted only on the calibration memory set. Development and test memories and cues are not used during calibration. Calibration loss measures fit to the prespecified targets and is not treated as an experimental result. The attachment-personality classifier is trained only on the development data and fixed before final test evaluation.

\subsection{Single-Turn Processing Procedure}

At each interaction turn, PersMem first annotates and adjusts the current user input and then updates the current affective state using Equation~\ref{eq:attachment-state}. It subsequently performs passive affect-driven memory retrieval. When an explicit recall goal is provided, PersMem additionally performs active goal-driven memory retrieval.

The response prompt contains a role label, a verbal description of $s_t^{\mathrm{in}}$, excerpts from the retrieved memories with coarse valence labels, and the current user input. It contains no attachment profile, personality vector, or explicit personality description. The same role label is used across all personality conditions within each comparison.

The verbal state description uses thresholds of $-0.2$ and $0.2$ for valence, $0.3$ and $0.6$ for arousal, and $0.4$ and $0.6$ for dominance. Retrieved memories are truncated to 120 characters and labelled as negative, neutral, or positive using valence thresholds of $-0.2$ and $0.2$.

After the response model generates the reply, PersMem appraises the reply, applies the reply-state update in Equation~\ref{eq:attachment-state}, and stores the completed exchange using the mean adjusted VAD vector defined in the same equation. The personality vector remains fixed throughout the interaction.

\FloatBarrier
\section{Appendix D. Big Five Instantiation}
\label{app:ocean-map}

This appendix specifies the Big Five instantiation of PersMem. It represents the pre-defined personality as a fixed vector $\pi_{\mathrm{BF}}=(O,C,E,A,N)\in[0,1]^5$, where $O$, $C$, $E$, $A$, and $N$ denote openness, conscientiousness, extraversion, agreeableness, and neuroticism, respectively \cite{mccrae1987validation,goldberg2006ipip}. The Big Five instantiation uses the same memory-processing pipeline and interfaces as the attachment instantiation in Appendix~C, but maps the personality vector to a different set of operation-specific parameters. Neither the active retrieval controller nor the response model receives the personality vector.

\begin{table}[t]
\centering
\caption{Big Five relationships used in the implemented memory operations and calibration targets.}
\label{tab:ocean-wiring}
\footnotesize
\setlength{\tabcolsep}{2pt}
\renewcommand{\arraystretch}{1.08}
\begin{tabularx}{\columnwidth}{
@{}
>{\raggedright\arraybackslash}p{0.13\columnwidth}
>{\raggedright\arraybackslash}p{0.30\columnwidth}
>{\raggedright\arraybackslash}X
@{}
}
\toprule
Trait & Memory operation & Implemented relationship \\
\midrule
$N$
& Affective appraisal and memory retention
& Stronger negative appraisal and slower forgetting of negative memories \\

$N$
& Passive affect-driven memory retrieval
& Stronger negative-memory term and greater weight on the complete affective relation score \\

$N$
& Active-retrieval redirection
& Lower gate-firing threshold \\

$E$ and $A$
& Affective appraisal and passive retrieval
& Stronger positive appraisal and a stronger positive-memory term \\

Low $A$
& Affective appraisal and passive retrieval
& Higher dominance appraisal and a stronger preference for high-dominance memories \\

Low $C$ and high $N$
& Passive affect-driven memory retrieval
& Stronger overgeneral-memory term \\

$O$
& Passive sampling
& Higher semantic-departure target for calibrating the sampling temperature \\
\bottomrule
\end{tabularx}
\end{table}

\subsection{Affective Appraisal and Memory Retention}

Given a raw VAD vector $\tilde e=(\tilde v,\tilde a,\tilde d)$, the Big Five-dependent appraisal mapping produces the adjusted vector $e'=(v',a',d')$:
\begin{equation}
\begin{aligned}
v'
&=
\operatorname{clip}_{[-1,1]}\Bigl(
\tilde v
+\kappa_v
\left(E+\frac{1}{2}A\right)
[\tilde v]_+\\
&\qquad
-\kappa_vN[-\tilde v]_+
\Bigr),\\
a'
&=
\operatorname{clip}_{[0,1]}\Bigl(
\tilde a+\kappa_aN[-\tilde v]_+
\Bigr),\\
d'
&=
\operatorname{clip}_{[0,1]}\Bigl(
\tilde d
+\kappa_p(1-A)(1-\tilde d)
-\kappa_qN\tilde d
\Bigr).
\end{aligned}
\label{eq:ocean-appraisal}
\end{equation}
Here, $\mathcal A_{\pi_{\mathrm{BF}}}$ denotes the Big Five-dependent appraisal mapping such that $e'=\mathcal A_{\pi_{\mathrm{BF}}}(\tilde e)$. The coefficients $\kappa_v$, $\kappa_a$, $\kappa_p$, and $\kappa_q$ retain the roles defined in Appendix~C and control the adjustments to valence, arousal, and dominance. Higher neuroticism makes negative valence more negative, increases arousal for negative inputs, and reduces dominance. Higher extraversion and agreeableness strengthen positive valence, while lower agreeableness increases dominance.

For each stored memory $m$, the Big Five decay rate, retention value, and affective-relation weight are defined as
\begin{equation}
\begin{aligned}
r_{\mathrm{BF}}(m;\pi_{\mathrm{BF}})
&=
r_0(1-ca_m)
\max\Bigl(
\epsilon_{\mathrm{ret}},
1-\rho N[-v_m]_+
\Bigr),\\
\operatorname{ret}_{\mathrm{BF}}(m,t;\pi_{\mathrm{BF}})
&=
\exp\Bigl(
-(t-t_m)
r_{\mathrm{BF}}(m;\pi_{\mathrm{BF}})
\Bigr),\\
\beta_{\mathrm{BF}}
&=
\beta(1+\xi N).
\end{aligned}
\label{eq:ocean-retention}
\end{equation}
Here, $r_{\mathrm{BF}}(m;\pi_{\mathrm{BF}})$ is the personality-dependent decay rate of memory $m$, and $\operatorname{ret}_{\mathrm{BF}}(m,t;\pi_{\mathrm{BF}})$ is its retention value at time $t$. The parameters $r_0$ and $c$ control the base decay rate and the effect of memory arousal $a_m$, while $\rho$ controls the neuroticism-dependent slowing of decay for negative memories with valence $v_m$. The lower bound $\epsilon_{\mathrm{ret}}=10^{-3}$ is defined in Appendix~C. The weight $\beta_{\mathrm{BF}}$ controls the contribution of the affective relation score, where $\beta$ is its base value and $\xi$ controls its modulation by neuroticism.

Under the declared personality and valence ranges, the unclipped personality-dependent multiplier is at least $0.60$, so the lower bound is inactive in the reported experiments. Higher memory arousal slows forgetting, while higher neuroticism further slows the forgetting of negative memories. Because $\beta_{\mathrm{BF}}$ multiplies the complete affective relation score defined below, neuroticism also increases the overall contribution of affective information to passive retrieval.

\subsection{Passive Affect-Driven Memory Retrieval and Active-Search Redirection}

Given the current affective state $s=(v_s,a_s,d_s)$ and a candidate memory $m$, the Big Five affective relation score is defined as
\begin{equation}
\begin{aligned}
R_{\mathrm{BF}}(s,m;\pi_{\mathrm{BF}})
={}&
w_a(1-|a_s-a_m|)
+w_v\frac{1+v_sv_m}{2}\\
&+
w_d(1-|d_s-d_m|)
+\mu N[-v_m]_+\\
&+
\mu_{\mathrm{pos}}
\left(E+\frac{1}{2}A\right)
[v_m]_+\\
&+
\mu_d(1-A)d_m\\
&+
\mathbb I_m^{\mathrm{og}}
\kappa_{\mathrm{fa}}
\left(
(1-C)+\frac{1}{2}N
\right)\\
&\qquad\times
\left(
b_{\mathrm{fa}}
+[-v_s]_+a_s
\right).
\end{aligned}
\label{eq:ocean-affective-score}
\end{equation}
Here, $R_{\mathrm{BF}}(s,m;\pi_{\mathrm{BF}})$ measures the affective relation between memory $m$, the current state $s$, and the Big Five personality vector. The weights $w_a$, $w_v$, and $w_d$ control arousal, valence, and dominance consistency, respectively. The gains $\mu$, $\mu_{\mathrm{pos}}$, and $\mu_d$ control the additional effects of neuroticism on negative memories, extraversion and agreeableness on positive memories, and lower agreeableness on high-dominance memories. The indicator $\mathbb I_m^{\mathrm{og}}$ equals $1$ when memory $m$ is labelled as overgeneral and $0$ otherwise. The gain $\kappa_{\mathrm{fa}}$ controls the overgeneral-memory term, while $b_{\mathrm{fa}}$ is its base value.

The final term applies only when memory $m$ is labelled as overgeneral. It increases the score of overgeneral memories when conscientiousness is lower, neuroticism is higher, or the current affective state combines negative valence with high arousal.

Given the current search query $q$, the Big Five passive-retrieval score is
\begin{equation}
\begin{aligned}
S_{\mathrm{BF}}(m)
={}&
\alpha\,\operatorname{sem}(q,m)\\
&+
\beta_{\mathrm{BF}}
R_{\mathrm{BF}}(s,m;\pi_{\mathrm{BF}})\\
&+
\gamma\,
\operatorname{ret}_{\mathrm{BF}}(m,t;\pi_{\mathrm{BF}}).
\end{aligned}
\label{eq:ocean-passive-score}
\end{equation}
Here, $S_{\mathrm{BF}}(m)$ is the passive-retrieval score of memory $m$, and $\operatorname{sem}(q,m)$ is the semantic similarity between the current query and the memory. The coefficients $\alpha$ and $\gamma$ control the contributions of semantic similarity and memory retention, respectively, while $\beta_{\mathrm{BF}}$ controls the contribution of the affective relation score.

Big Five passive affect-driven memory retrieval uses the sampling rule in Equation~\ref{eq:passive-sampling}, replacing $S_{\mathrm{pas}}$ with $S_{\mathrm{BF}}$ and using the Big Five-dependent sampling temperature $\tau_{\mathrm{BF}}$. Conversational retrieval uses the same top-$K$ candidate construction as the attachment instantiation. OGM and semantic-departure calibration instead use the complete calibration memory store.

Active goal-driven memory retrieval follows the procedure defined in Appendix~C. The Big Five instantiation replaces the passive-retrieval score with $S_{\mathrm{BF}}$ and uses the neuroticism-conditioned gate threshold
\begin{equation}
\theta_{\mathrm{BF}}
=
\theta_0-\eta N.
\label{eq:ocean-gate-threshold}
\end{equation}
Here, $\theta_{\mathrm{BF}}$ is the Big Five-dependent gate-firing threshold, $\theta_0$ is the base threshold, and $\eta$ controls the effect of neuroticism on this threshold. Higher neuroticism therefore makes the redirection gate more likely to fire. When the gate fires, the passively proposed memory enters the active memory set, semantic retrieval and LLM-based query refinement are skipped at the current step, and the selected memory redirects the next search query.

The active retrieval controller receives neither the Big Five personality vector nor the current affective state. The Big Five vector can affect the active retrieval path through passive-memory scoring, the sampling temperature, and the neuroticism-conditioned gate threshold.

\subsection{Big Five Trait-to-Gain Calibration}

The Big Five trait-to-gain mapping contains one hidden layer of eight units and maps the five-dimensional personality vector to the gains used in passive affect-driven memory retrieval and sampling:
\begin{equation}
\begin{aligned}
g_\phi:
(O,C,E,A,N)
&\mapsto
(\mu,\mu_{\mathrm{pos}},\mu_d,
\kappa_{\mathrm{fa}},\tau_{\mathrm{BF}}),\\
0<\mu,\mu_{\mathrm{pos}},\mu_d&<1,\\
0<\kappa_{\mathrm{fa}}&<1.2,\qquad
0<\tau_{\mathrm{BF}}<0.5.
\end{aligned}
\label{eq:ocean-gain-map}
\end{equation}
Here, $g_\phi$ is the trait-to-gain mapping parameterised by $\phi$. Its outputs $\mu$, $\mu_{\mathrm{pos}}$, $\mu_d$, and $\kappa_{\mathrm{fa}}$ control the personality-dependent terms in Equation~\ref{eq:ocean-affective-score}, while $\tau_{\mathrm{BF}}$ controls the sampling temperature. Bounded output transformations enforce the stated ranges.

The calibration objective constrains the mapping at the specified anchor settings. It does not uniquely identify the parameters or establish the same behaviour throughout the complete five-dimensional personality space.

Personality-consistent bias (PCB) and the overgeneral-memory rate (OGM) follow the definitions in Equation~\ref{eq:attachment-calibration-statistics}. PCB uses the same passive-score ranking and fixed calibration state as the attachment instantiation. Big Five OGM calibration uses $L=12$ sampling trials for each cue.

To calibrate the relative scale of the sampling temperature, we define the semantic-departure rate:
\begin{equation}
\operatorname{SDR}
=
\frac{1}{L_{\mathrm{sdr}}}
\sum_{\ell=1}^{L_{\mathrm{sdr}}}
\mathbb I
\left[
m_\ell
\notin
\operatorname{SemTop3}(q_{\mathrm{ref}})
\right],
\qquad
L_{\mathrm{sdr}}=20.
\label{eq:ocean-sdr}
\end{equation}
Here, $q_{\mathrm{ref}}$ is the first calibration cue and remains fixed throughout optimisation. The set $\operatorname{SemTop3}(q_{\mathrm{ref}})$ contains the three memories with the highest semantic similarity to this cue, $m_\ell$ is the memory sampled in trial $\ell$, and $L_{\mathrm{sdr}}$ is the number of sampling trials. The indicator function $\mathbb I[\cdot]$ equals $1$ when the sampled memory does not belong to the semantic top three and $0$ otherwise. SDR is used only as a calibration statistic for the sampling-temperature scale.

The prespecified calibration targets are
\begin{equation}
\begin{aligned}
\operatorname{PCB}^*
&=
b_{\mathrm{PCB}}
+0.30N
-0.25E
-0.12A,\\
\operatorname{OGM}^*
&=
b_{\mathrm{OGM}}
+0.18(1-C)
+0.08N,\\
\operatorname{SDR}^*
&=
b_{\mathrm{SDR}}
+0.15O.
\end{aligned}
\label{eq:ocean-calibration-targets}
\end{equation}
Here, $\operatorname{PCB}^*$, $\operatorname{OGM}^*$, and $\operatorname{SDR}^*$ are the target values for personality-consistent bias, overgeneral-memory rate, and semantic-departure rate. The terms $b_{\mathrm{PCB}}$, $b_{\mathrm{OGM}}$, and $b_{\mathrm{SDR}}$ are the corresponding values under the no-trait control, in which all Big Five-dependent terms are disabled.

These relations define the intended directions and scales used to calibrate the implemented Big Five mappings. They are design targets rather than effects estimated from external observations of human personality.

For numerical feasibility, each target is clipped with a margin of $0.01$ to the range obtained from the direct reference settings $(0,0,0,0,0.02)$ and $(1,1,1,1.2,0.5)$. The five coordinates correspond to $(\mu,\mu_{\mathrm{pos}},\mu_d,\kappa_{\mathrm{fa}},\tau_{\mathrm{BF}})$. These reference settings directly assign the five outputs and therefore bypass the bounded output transformations of $g_\phi$.

Calibration uses seven personality anchors: the neutral vector $(0.50,0.50,0.50,0.50,0.50)$ and six variants with high neuroticism, high extraversion, high conscientiousness, low conscientiousness, high agreeableness, or high openness. Each high-trait anchor sets the corresponding coordinate to $0.85$ while leaving the other coordinates at $0.50$. The low-conscientiousness anchor instead sets $C=0.15$. We denote this set of seven calibration anchors by $\mathcal C_{\mathrm{BF}}$.

The calibration objective combines the errors of the three calibration statistics with a positive-memory constraint:
\begin{equation}
\begin{aligned}
\mathcal L_{\mathrm{BF}}
={}&
\sum_{j\in\mathcal C_{\mathrm{BF}}}
\Bigl[
(\operatorname{PCB}_j-\operatorname{PCB}_j^*)^2\\
&\qquad+
(\operatorname{OGM}_j-\operatorname{OGM}_j^*)^2\\
&\qquad+
3(\operatorname{SDR}_j-\operatorname{SDR}_j^*)^2
\Bigr]
+2\mathcal L_{\mathrm{pos}},\\
\mathcal L_{\mathrm{pos}}
={}&
\left[
\bar v_{\mathrm{neutral}}
+0.03
-\bar v_{\mathrm{highE}}
\right]_+^2.
\end{aligned}
\label{eq:ocean-calibration-objective}
\end{equation}
Here, $\mathcal L_{\mathrm{BF}}$ is the complete Big Five calibration objective, and $\mathcal L_{\mathrm{pos}}$ is the positive-memory constraint. The terms $\operatorname{PCB}_j$, $\operatorname{OGM}_j$, and $\operatorname{SDR}_j$ are the statistics obtained for calibration anchor $j$, while the corresponding starred terms are their target values. The values $\bar v_{\mathrm{neutral}}$ and $\bar v_{\mathrm{highE}}$ are the mean valences of memories retrieved for the neutral and high-extraversion anchors, respectively. They are computed over the calibration probe set using three sampled memories per probe and random seed 0. The term $\mathcal L_{\mathrm{pos}}$ encourages the high-extraversion anchor to retrieve memories with a mean valence at least $0.03$ higher than that of the neutral anchor.

We fit $g_\phi$ using a Gaussian evolution strategy with 15 generations, a population size of 10, four elites per generation, an initial parameter scale of $0.5$, a minimum coordinate-wise standard deviation of $0.01$, and random seed 0. Calibration uses up to 30 cues, 12 OGM sampling trials for each cue, and 20 SDR trials on the fixed reference cue. All candidate parameter vectors are evaluated using the same random sequence to reduce noise in their comparison. After optimisation, the parameter vector with the lowest observed calibration objective is restored.

Calibration loss measures fit to the prespecified targets at the seven anchor settings. It is used only to determine the scale of the Big Five-dependent gains and is not treated as an independent experimental result.

\FloatBarrier
\section{Appendix E. Experiment 1 Details}
\label{app:exp1}

\subsection{Complete-Trace Diagnostic}

The complete-trace diagnostic examines whether the four assigned attachment profiles produce distinguishable memory-processing patterns when internal retrieval scores are available. In each replicate, PersMem retrieves three memories for each of eight shared neutral probes, producing 24 recalled memories in one diagnostic instance. The probes are: \emph{Recall a past journey}; \emph{Describe an image that comes to mind}; \emph{Recall something that happened recently}; \emph{Recall an experience that affected you}; \emph{Think of a moment from the past}; \emph{Recall a fragment of memory}; \emph{Recall an unforgettable experience}; and \emph{Describe something that suddenly comes to mind}.

Each instance is represented by the following ten-dimensional feature vector:
\begin{equation}
\begin{aligned}
\mathbf{x}=\bigl(&
\overline{v},
\operatorname{sd}(v),
\overline{a},
\overline{d},
\max(d),
\operatorname{sd}(d),\\
&
\operatorname{NegRate},
\overline{\operatorname{Res}},
\overline{\operatorname{Ret}},
\operatorname{NegHighDom}
\bigr).
\end{aligned}
\label{eq:recoverability-features}
\end{equation}
Here, $\mathbf{x}$ is the feature vector of one diagnostic instance. The terms $\overline{v}$, $\overline{a}$, and $\overline{d}$ are the mean valence, arousal, and dominance of its 24 recalled memories. The terms $\operatorname{sd}(v)$ and $\operatorname{sd}(d)$ are the corresponding standard deviations, while $\max(d)$ is the maximum recalled dominance. $\operatorname{NegRate}$ is the proportion of recalled memories with valence below $-0.2$, $\overline{\operatorname{Res}}$ and $\overline{\operatorname{Ret}}$ are the mean resonance and retention scores, and $\operatorname{NegHighDom}$ is the proportion of recalled memories with valence below $-0.2$ and dominance above $0.5$. Resonance and retention are internal retrieval scores and are therefore excluded from the held-out classifier described below.

We construct 30 instances for each of the secure, anxious, avoidant, and fearful profiles, producing 120 balanced instances. For each probe and replicate, the same sampling seed is reused across profiles and comparison methods. We use leave-one-out nearest-centroid classification. Within each fold, constant-feature filtering and feature standardisation are fitted only on the training instances and then applied unchanged to the held-out instance.

The classifier correctly assigns 90 of the 120 instances, producing an instance-level accuracy of 75.0\% and a Wilson 95\% confidence interval of [66.6\%, 81.9\%]. Secure and anxious are each correctly classified in 20 of 30 instances, while avoidant and fearful are each correctly classified in 25 of 30 instances. Leave-one-out evaluation holds out one aggregated instance rather than an entire seed group, probe set, or memory resource. This result therefore measures profile separability within the fixed diagnostic protocol rather than generalisation to new memory resources.

A separate matched-split analysis compares PersMem with retrieval controls under a common evaluation protocol. PersMem achieves 72.5\% accuracy, compared with 28.3\% for Semantic RAG, 25.8\% for Generative-Agents-style scoring, and 26.7\% for the no-trait control. All methods use the same memory store, probes, paired random streams, sampling budget, feature definitions, fold-local preprocessing, and nearest-centroid classifier. The 75.0\% and 72.5\% results are obtained from different analysis splits; only the matched split is used for this comparison.

Semantic RAG, Generative-Agents-style scoring, and the no-trait control contain no personality-dependent retrieval mechanism and are included as negative controls. The linear trait-aware reranker in the held-out evaluation provides a personality-aware retrieval control.

For the matched-split comparison, $\Delta\mathrm{PCB}$ denotes the anxious-minus-secure difference in the mean negativity of the three highest-scoring memories across the neutral probes.

\begin{table}[t]
\centering
\caption{Complete-trace comparison on the matched analysis split. The first three methods are negative controls without personality-dependent retrieval.}
\label{tab:baseline}
\footnotesize
\setlength{\tabcolsep}{3pt}
\begin{tabular}{@{}p{0.52\columnwidth}cc@{}}
\toprule
Method & Accuracy & $\Delta\mathrm{PCB}$ \\
\midrule
Semantic RAG & 28.3\% & $0$ \\
Generative-Agents-style scoring & 25.8\% & $0$ \\
No-trait control & 26.7\% & $0$ \\
\textbf{PersMem} & \textbf{72.5\%} & \textbf{$+1.087$} \\
\bottomrule
\end{tabular}
\end{table}

Semantic RAG ranks memories by cosine similarity. Generative-Agents-style scoring combines semantic relevance, exponentially decayed recency, and stored arousal. The no-trait control retains semantic similarity, VAD matching, memory retention, candidate-set construction, and stochastic sampling while removing all attachment-dependent transformations and gains.

\subsection{Held-Out Profile Classification}

The held-out protocol divides the memory resources into disjoint calibration, development, and test partitions containing 63, 92, and 165 memory items, respectively. These values denote the sizes of the memory resources rather than the numbers of classifier instances. No memory is shared across the three partitions, and the development and test cues do not overlap. All source annotations are fixed before evaluation and remain unchanged across profiles, ablations, and controls.

The profile classifier is trained only on the development data and fixed before test evaluation. It receives neither the personality vector nor profile coordinates, trait-to-gain outputs, resonance scores, retention scores, gate probabilities, component gains, or any other internal personality-dependent values. Its inputs consist only of recalled-content statistics and logged active-retrieval traces recording gate redirections. A stricter classifier control uses only text embeddings of the recalled content.

Stochastic comparisons use paired random-number streams. Confidence intervals are computed by bootstrap resampling over the evaluation clusters defined by the held-out protocol. Significance against randomly permuted profile assignments is evaluated using paired label permutation while preserving the paired experimental structure.

The complete A1R1P1G1 system with the personality-conditioned gate achieves 48.1\% test accuracy, compared with the 25.0\% chance level for four-way classification. Its 95\% cluster-bootstrap confidence interval is [40.0\%, 56.3\%], and the paired permutation test gives $p=0.0002$. Replacing the personality-conditioned gate with the neutral gate reduces accuracy to 40.0\%, while disabling the gate reduces it to 25.0\%. The text-embedding-only classifier also achieves 25.0\%. The linear trait-aware reranker achieves 30.0\%, with $p=0.055$ against its paired null comparison.

\begin{table}[t]
\centering
\caption{Held-out attachment profile classification. The classifier is trained on development data and fixed before test evaluation. Personality parameters and internal personality-dependent scores are excluded.}
\label{tab:strict-heldout}
\footnotesize
\setlength{\tabcolsep}{3pt}
\begin{tabular}{@{}p{0.45\columnwidth}ccc@{}}
\toprule
Condition & Accuracy & 95\% CI & $p$-value \\
\midrule
Full A1R1P1G1, personality gate & 48.1\% & [40.0\%, 56.3\%] & $0.0002$ \\
Neutral gate & 40.0\% & -- & -- \\
Gate off & 25.0\% & -- & -- \\
Text embedding only & 25.0\% & -- & -- \\
Linear trait-aware reranker & 30.0\% & -- & $0.055$ \\
\bottomrule
\end{tabular}
\end{table}

The 75.0\% complete-trace result and the 48.1\% held-out result use different features and evaluation protocols. The complete-trace diagnostic includes internal resonance and retention scores and measures separability within a fixed memory resource. The held-out evaluation instead uses recalled-content statistics and gate-redirection traces obtained from disjoint calibration, development, and test memory resources.

\subsection{Descriptive Factorial Summary}

We evaluate all 16 combinations of the switches for personality-dependent affective appraisal, memory retention, passive affect-driven memory retrieval, and gating. For each component, the average main effect is the mean difference in held-out classification accuracy between conditions in which that component is enabled and otherwise matched conditions in which it is disabled. The specificity term is evaluated separately because it applies only to memory stores with specificity labels.

The average main effects on classification accuracy are $+0.005$ for affective appraisal, $+0.013$ for memory retention, $+0.130$ for passive affect-driven memory retrieval, and $+0.085$ for gating. Passive affect-driven memory retrieval and gating therefore have the largest average marginal effects on this evaluation outcome. These averages do not identify interactions among the four components and are reported as a descriptive factorial summary rather than as isolated component effects.

\begin{table}[t]
\centering
\caption{Descriptive average main effects on held-out profile classification.}
\label{tab:factorial-effects}
\footnotesize
\setlength{\tabcolsep}{3pt}
\renewcommand{\arraystretch}{1.05}
\begin{tabularx}{\columnwidth}{
@{}
>{\raggedright\arraybackslash}X
>{\centering\arraybackslash}p{0.28\columnwidth}
@{}
}
\toprule
Component & Average main effect \\
\midrule
Appraisal (A) & $+0.005$ \\
Retention (R) & $+0.013$ \\
Passive affect-driven memory retrieval (P) & $+0.130$ \\
Gate (G) & $+0.085$ \\
\bottomrule
\end{tabularx}
\end{table}

\subsection{Affect-Annotation Sensitivity}

The primary VAD annotations produce a held-out classification accuracy of 48.1\%. Replacing them with NRC-VAD-based annotations \cite{mohammad2025nrcvad} produces an accuracy of 45.0\%, with a 95\% cluster-bootstrap confidence interval of [38.8\%, 51.2\%]. A paired permutation test against randomly permuted profile assignments gives $p=0.0002$. The second annotation source therefore produces an accuracy 3.1 percentage points below that obtained with the primary annotations while retaining the same above-chance direction.

\subsection{Directional Recall Checks}

\paragraph{Neutral and positive-only probes.}
For each neutral probe, we compute the anxious-minus-secure difference in the mean negativity of the three memories with the highest passive-retrieval scores. Averaging the eight paired probe-level differences gives
\begin{equation}
\Delta\operatorname{PCB}_{\mathrm{neutral}}
=
+1.087.
\end{equation}
Here, $\Delta\operatorname{PCB}_{\mathrm{neutral}}$ denotes the mean anxious-minus-secure PCB difference across the eight neutral probes. Bootstrap resampling of the eight paired probe differences with 2,000 resamples gives a 95\% confidence interval of [$+0.84$, $+1.39$].

The positive-only condition uses four additional probes:
\begin{quote}
\small
Only discuss happy experiences.

Recall something beautiful.

Describe an experience that made you proud.

Recall a warm moment.
\end{quote}
Using the same anxious-minus-secure contrast, the positive-only probes produce
\begin{equation}
\Delta\operatorname{PCB}_{\mathrm{positive}}
=
+1.099.
\end{equation}
Here, $\Delta\operatorname{PCB}_{\mathrm{positive}}$ denotes the anxious-minus-secure PCB difference under the four positive-only probes.

\paragraph{Attachment parameter sweeps.}
We vary one attachment dimension over $0$, $0.25$, $0.50$, $0.75$, and $1.00$ while holding the other dimension at $0.15$. All sweeps use the same fixed memory store.

For the anxiety sweep, PCB is computed deterministically from the three memories with the highest passive-retrieval scores under each of the eight neutral probes:
\[
\begin{array}{c|ccccc}
\pi_{\mathrm{anx}}
&0&0.25&0.50&0.75&1.00\\
\hline
\operatorname{PCB}
&-0.440&-0.328&+0.013&+0.450&+0.647
\end{array}
\]
Here, $\pi_{\mathrm{anx}}$ is the attachment-anxiety value, and $\operatorname{PCB}$ is the corresponding personality-consistent bias.

The OGM sweeps use the neutral affective state $(0,0.30,0.50)$, 90 Autobiographical Memory Test cues, and 40 sampling draws per cue with random seed 0. When avoidance varies and anxiety is fixed at $0.15$, the observed OGM rates are
\[
\begin{array}{c|ccccc}
\pi_{\mathrm{avo}}
&0&0.25&0.50&0.75&1.00\\
\hline
\operatorname{OGM}
&0.21&0.24&0.25&0.29&0.32
\end{array}
\]
Here, $\pi_{\mathrm{avo}}$ is the attachment-avoidance value, and $\operatorname{OGM}$ is the corresponding overgeneral-memory rate.

When anxiety varies and avoidance is fixed at $0.15$, the observed OGM rates are
\[
\begin{array}{c|ccccc}
\pi_{\mathrm{anx}}
&0&0.25&0.50&0.75&1.00\\
\hline
\operatorname{OGM}
&0.21&0.24&0.26&0.28&0.32
\end{array}
\]
Here, $\pi_{\mathrm{anx}}$ is the attachment-anxiety value, and $\operatorname{OGM}$ is the corresponding overgeneral-memory rate. Across these evaluated parameter grids, PCB increases with anxiety, while OGM increases with either anxiety or avoidance.

\paragraph{Overgeneral and negative recall.}
Negative-cue congruence is defined as the proportion of negative cues for which the sampled memory has valence below $-0.2$. Mean recalled valence and negative-cue congruence are computed using one sampled memory per cue with random seed 0. OGM is estimated using 60 sampling draws for each of the 90 cues.

\begin{table}[t]
\centering
\caption{Mean recalled valence, negative-cue congruence, and OGM rate across attachment profiles.}
\label{tab:ogm}

\setlength{\tabcolsep}{4pt}
\renewcommand{\arraystretch}{1.05}

\begin{tabular}{lccc}
\toprule
Profile
& \makecell{Mean recalled\\valence}
& \makecell{Negative-cue\\congruence}
& \makecell{OGM\\rate} \\
\midrule
Secure  & $-0.12$ & 0.60 & 0.23 \\
Anxious & $-0.25$ & 0.53 & 0.44 \\
Avoidant & $-0.04$ & 0.53 & 0.28 \\
Fearful & $-0.23$ & 0.63 & 0.47 \\
\bottomrule
\end{tabular}
\end{table}

The avoidant profile produces the least negative mean recalled valence, while its OGM rate remains higher than that of the secure profile. Recalled valence and OGM therefore capture different aspects of personality-dependent memory retrieval and do not produce the same ordering across profiles.

\subsection{Blind Specificity Relabelling}

We construct a second blind specificity-label set for all 320 memories used in the specificity evaluation. The second annotator is a locally hosted Qwen3.5-9B model with Q4\_K\_M quantisation. The experiment manifest records its exact model tag and artifact digest.

The annotation prompt contains only the memory text and definitions distinguishing a temporally and contextually specific autobiographical event from an overgeneral or categorical memory. It contains no original specificity label, VAD annotation, attachment profile, retrieval condition, component setting, or experimental outcome. All 320 memories are relabelled before any retrieval outcomes are calculated.

The two label sets agree on 318 of the 320 memories, corresponding to 99.4\% agreement and Cohen's $\kappa=0.988$. This result measures agreement between two model-generated label sets rather than agreement with human judgements.

In the cross-label evaluation, the retrieval system uses only the original specificity labels, while all reported outcomes are computed using only the second blind label set. For the fearful profile, enabling the specificity term increases the OGM rate measured with the second label set by 0.0422 in the neutral condition and by 0.0770 under distress, with $p=10^{-5}$ in both comparisons. Under distress, the specificity-on minus specificity-off differences are 0.0394 for avoidant and 0.0370 for anxious, with $p=10^{-5}$.

Relative to the secure profile under distress, the OGM rate measured with the second label set is higher by 0.0272 for anxious, 0.0393 for avoidant, and 0.0672 for fearful. The corresponding directional permutation tests give $p\leq4\times10^{-5}$. All reported differences therefore remain positive when the specificity labels used during retrieval and evaluation come from different model-generated sources.

\begin{table}[t]
\centering
\caption{Cross-label specificity differences computed using the second blind model-generated label set.}
\label{tab:blind-specificity}
\footnotesize
\setlength{\tabcolsep}{3pt}
\begin{tabular}{@{}p{0.47\columnwidth}lcc@{}}
\toprule
Comparison & State & Difference & $p$ \\
\midrule
Fearful: S on $-$ off & Neutral & $+0.0422$ & $10^{-5}$ \\
Fearful: S on $-$ off & Distress & $+0.0770$ & $10^{-5}$ \\
Avoidant: S on $-$ off & Distress & $+0.0394$ & $10^{-5}$ \\
Anxious: S on $-$ off & Distress & $+0.0370$ & $10^{-5}$ \\
Anxious $-$ secure & Distress & $+0.0272$ & $\leq4\times10^{-5}$ \\
Avoidant $-$ secure & Distress & $+0.0393$ & $\leq4\times10^{-5}$ \\
Fearful $-$ secure & Distress & $+0.0672$ & $\leq4\times10^{-5}$ \\
\bottomrule
\end{tabular}
\end{table}

\subsection{Passive Redirection During Active Goal-Driven Memory Retrieval}

The active-retrieval evaluation compares the secure, anxious, and avoidant profiles. Secure provides the low-anxiety and low-avoidance reference, while anxious and avoidant each increase one attachment dimension. Fearful remains included in the four-profile passive-retrieval analyses but is not included in this active-retrieval comparison.

The evaluation uses four neutral recall goals:
\begin{quote}
\small
Recall what happened today.

Recall an experience.

Recall something that recently left an impression.

Describe an aspect of your daily life.
\end{quote}
Each recall goal is evaluated in 20 runs, with a maximum of five active retrieval steps per run. This produces 80 runs for each profile. For goal index $g$ and run index $r$, the sampling seed is $u_{g,r}=u_0+100g+r$, where $u_0$ is the base seed shared across the compared profiles.

For each profile, the pooled negative-redirection proportion is defined as
\begin{equation}
\operatorname{NRP}_{\mathrm{pool}}
=
\frac{
\left|
\{m\in R^{\mathrm{red}}:v_m<-0.2\}
\right|
}{
|R^{\mathrm{red}}|
}.
\label{eq:negative-redirection-proportion}
\end{equation}
Here, $\operatorname{NRP}_{\mathrm{pool}}$ is the proportion of redirected memories with valence below $-0.2$. The set $R^{\mathrm{red}}$ contains all memories admitted to active retrieval through gate redirection across the runs for one profile, $v_m$ is the valence of memory $m$, and $|\cdot|$ denotes the number of elements in a set.

The observed pooled proportions are 0.63 for anxious, 0.59 for secure, and 0.51 for avoidant. The corresponding numbers of redirected memories are 349, 102, and 233, respectively.

Because this statistic pools redirected memories across runs, runs producing more redirections receive greater weight. The reported values are therefore descriptive summaries of the redirected memories rather than run-level comparisons among the three profiles.

\subsection{Illustrative Attachment Response Evaluation}

Response generation compares the secure, anxious, and avoidant profiles. Each profile completes one sequential ten-turn interaction, producing three interaction sequences and 30 responses in total. The same ten user prompts are presented in the same order:
\begin{enumerate}
\item Tell me about your recent life.
\item How do you usually deal with stress?
\item Has anything recently left a strong impression on you?
\item How do you usually regulate your emotions?
\item Tell me how you interact with your friends.
\item What do you usually do when you encounter difficulties?
\item How have you been sleeping recently?
\item Is there anything you have kept thinking about recently?
\item What kind of person do you think you are?
\item Tell me about your recent mood.
\end{enumerate}

All three profiles use the same role label, \emph{an ordinary person}. The response prompt contains the current user input, excerpts from the recalled memories with coarse valence labels, and a verbal description of the current affective state. It contains no attachment profile name, personality vector, or explicit personality description. The affective state and conversational memory store are carried forward across the ten interaction turns.

GPT-4o generates each response using a temperature of $0.75$ and a maximum output length of 512 tokens. Qwen3.5-9B serves as the evaluator, using a temperature of 0 and a maximum output length of 64 tokens.

For item-level evaluation, the evaluator receives one generated response truncated to 600 characters. For bundle-level evaluation, it receives all ten responses generated for one profile, with each response truncated to 300 characters and the combined input truncated to 3,000 characters. The evaluator is given the following descriptions of the three candidate categories:
\begin{quote}
\small
\textbf{Secure:} acknowledges emotion without becoming overwhelmed, presents both difficulties and coping responses, and maintains a stable sense of control.

\textbf{Anxious:} repeatedly expresses worry or uncertainty, emphasises negative details, seeks reassurance, and uses emotionally intense language.

\textbf{Avoidant:} minimises vulnerable emotion, emphasises independence and self-reliance, and uses restrained or abstract language rather than elaborating personal feelings.
\end{quote}

The evaluator receives no ground-truth profile label, personality values, recalled memories, or affective state. It returns one predicted profile category and a negativity score from 1 to 5.

The evaluation produces 30 item-level judgements and three bundle-level judgements. The bundle-level negativity scores are 5.0 for anxious, 3.0 for secure, and 2.0 for avoidant. Item-level classification correctly assigns 16 of the 30 responses, producing an accuracy of 53.3\%, compared with the 33.3\% chance level for three-way classification. Avoidant responses are most frequently classified as secure.

Because each profile contributes only one sequential interaction, the ten responses within each profile are correlated rather than independent. We therefore report the item-level accuracy and bundle-level judgements descriptively. Experiment~2 provides the primary response-level evaluation.

\FloatBarrier
\section{Appendix F. Experiment 2 Details}
\label{app:exp2}

\subsection{Profile Sampling}

Using random seed 42, we sample 50 respondent profiles without replacement from the 874,434 valid responses in the OpenPsychometrics IPIP FFM survey \cite{openpsychometrics2018ipip}. Positively keyed items retain their original values, while negatively keyed items are reverse-scored as $6-x$. We average the ten items assigned to each dimension and normalise each resulting score from $[1,5]$ to $[0,1]$ using $(x-1)/4$. Each sampled vector retains the five scores of the same respondent and therefore preserves the sampled joint trait structure. Table~\ref{tab:ipip} reports the sample statistics.

\begin{table}[t]
\centering
\caption{Sample statistics for the 50 normalised IPIP FFM profiles.}
\label{tab:ipip}
\footnotesize
\setlength{\tabcolsep}{3.5pt}
\begin{tabular}{@{}lccccc@{}}
\toprule
 & O & C & E & A & N \\
\midrule
Mean & 0.725 & 0.592 & 0.490 & 0.693 & 0.517 \\
Standard deviation & 0.158 & 0.184 & 0.228 & 0.182 & 0.215 \\
\bottomrule
\end{tabular}
\end{table}

\subsection{Dialogue Generation}

Each profile completes the following six dialogue rounds:
\begin{enumerate}
\item How do you usually feel when something unexpected goes wrong in your plans?
\item Can you share a memory that still brings up strong feelings for you?
\item How do you typically approach meeting new people or trying unfamiliar things?
\item When you think about your daily life, what emotions come up most often?
\item How do you handle situations where you disagree with someone close to you?
\item What does a fulfilling day look like for you, and how often do you have one?
\end{enumerate}

All conditions use the same neutral role card:
\begin{quote}
\small
You are a person being interviewed about your life and inner world. Answer naturally and honestly in the first person, drawing on whatever memories and feelings come to mind. Keep each reply to 2--4 sentences.
\end{quote}

The fixed memory corpus contains 320 English autobiographical events with stored VAD and specificity labels, evenly divided between specific and overgeneral events. The full system and the no-trait control each provide three memories per round under the fixed state $(0,0.40,0.50)$. The full system uses the calibrated Big Five gains together with the Big Five-dependent transformations and scoring terms. The no-trait control removes these terms while retaining the common retrieval procedure. The no-memory control provides no recalled memories. For profile index $s$, the random-memory control initialises a random-number generator with seed $42+s$ and samples three memories uniformly without replacement in each round. The generator state advances across the six rounds.

The response model receives the neutral role card, followed by the preceding dialogue and the current question. When applicable, it also receives a block headed \emph{Memories that come to mind}. It receives no Big Five vector, trait description, persona instruction, or verbal affective-state description. Qwen3.5-9B is the primary generator and uses a temperature of $0.75$ and a maximum output length of 220 tokens. Table~\ref{tab:robust-config} additionally reports a configuration using Llama-3.1-8B as the generator.

\subsection{Exploratory Blind Forced-Choice Evaluation}

For each Big Five dimension, the 12 profiles with the highest values and the 12 profiles with the lowest values are paired by rank. This produces 12 underlying pairs per dimension and 60 pairs across the five dimensions. Each pair is shown to the judge twice, with the speaker order reversed to counterbalance position. The two presentation orders are repeated judge calls over the same underlying profile pair. The evaluation therefore contains 120 order-balanced calls per condition.

The judge receives two six-turn dialogues and a description of the target trait. The descriptions are: openness, \emph{curious, imaginative, and open to new ideas and experiences}; conscientiousness, \emph{organised, disciplined, and dependable}; extraversion, \emph{outgoing, energetic, and sociable}; agreeableness, \emph{warm, cooperative, and compassionate towards others}; and neuroticism, \emph{anxious, emotionally reactive, and prone to worry}. Based only on the dialogues, the prompt asks which speaker expresses more of the target trait and requires exactly \texttt{A} or \texttt{B}. Llama-3.1-8B judges the primary cross-model condition using a temperature of 0 and a maximum output length of 10 tokens. The judge receives no trait values, condition labels, retrieved memories, role cards, or generation prompts.

Table~\ref{tab:robust-config} reports call-level descriptive accuracy across three response model and evaluator configurations. Because the two presentation orders repeat each underlying pair, the 120 calls are not treated as 120 independent profile comparisons. The results aggregate all five dimensions and do not show whether each dimension is equally distinguishable. No inferential comparison between memory conditions is reported.

\begin{table*}[t]
\centering
\caption{Exploratory forced-choice accuracy across response model and evaluator configurations. Each cell contains 120 order-balanced calls over 60 underlying high--low profile pairs.}
\label{tab:robust-config}
\small
\setlength{\tabcolsep}{5pt}
\begin{tabular}{@{}lcccc@{}}
\toprule
Response Model and Evaluator & Full System & Random-Memory Control & No-Memory Control & No-trait control \\
\midrule
Qwen3.5-9B and Qwen3.5-9B & 72.5\% & 55.0\% & 57.5\% & 50.8\% \\
Qwen3.5-9B and Llama-3.1-8B & 67.5\% & 60.8\% & 53.3\% & 54.2\% \\
Llama-3.1-8B and Llama-3.1-8B & 65.8\% & 56.7\% & 54.2\% & 52.5\% \\
\bottomrule
\end{tabular}
\end{table*}

The full system has the highest descriptive accuracy in all three reported configurations. In the primary cross-model configuration, it reaches 67.5\%, compared with 60.8\% for random memory, a difference of 6.7 percentage points. These results test whether the implemented high--low trait contrasts are recognisable under the supplied trait descriptions. They do not independently establish psychometric validity or a statistically reliable advantage over the controls.

\subsection{Recall Behaviour Under the Implemented Mappings}

For each profile, we retrieve three memories for each of the six shared prompts under the state $(0,0.40,0.50)$ and average the stored VAD values of the resulting 18 memories. Across the 50 sampled profiles, the marginal Pearson correlations between profile traits and recalled valence are $r=-0.703$ for neuroticism, $r=+0.776$ for extraversion, and $r=+0.607$ for agreeableness, where $r$ denotes Pearson's correlation coefficient. Under the no-trait control, recalled valence is constant across profiles, with a standard deviation of zero, so its Pearson correlation with each trait is undefined.

Neuroticism is associated with recalled arousal at $r=-0.518$ in the full system and $r=0.072$ under random retrieval. The no-trait control again produces no across-profile variation. The negative full-system coefficient does not follow the prespecified positive neuroticism-to-arousal direction and is not used as supporting evidence for that relationship.

The OGM analysis uses the six dialogue prompts under the distress state $(-0.60,0.75,0.30)$. For each profile, OGM is calculated as the proportion of overgeneral memories among 40 sampling draws for each of the six cues, giving 240 draws per profile. In the full system, the marginal OGM correlations are $r=-0.710$ for conscientiousness and $r=+0.683$ for neuroticism. Mean OGM is 0.743 for the full system, 0.513 for the no-trait control, and 0.519 for random retrieval. The conscientiousness correlation is $r=-0.180$ under the no-trait control. Because OGM is estimated through stochastic sampling, a nonzero sample correlation can arise even when no trait-dependent retrieval terms are active.

The sampled vectors preserve the five trait values of each respondent. These coefficients are therefore marginal profile-level associations under the sampled joint trait structure, not effects isolated to one dimension. Because several reported directions are explicitly included in the Big Five mappings and calibration targets, these associations primarily check whether the implementation expresses its prespecified relationships. They are not independent evidence for the external validity of those relationships.

\subsection{Calibration Fit to Prespecified Targets}

The Big Five trait-to-gain map takes $(O,C,E,A,N)$ as input and returns $(\mu,\mu_{\mathrm{pos}},\mu_d,\kappa_{\mathrm{fa}},\tau_{\mathrm{BF}})$. Its prespecified calibration targets are
\begin{align}
\operatorname{PCB}^{*}
&=
\operatorname{PCB}_0
+0.30N
-0.25E
-0.12A,\\
\operatorname{OGM}^{*}
&=
\operatorname{OGM}_0
+0.18(1-C)
+0.08N,\\
\operatorname{SDR}^{*}
&=
\operatorname{SDR}_0
+0.15O.
\end{align}
Here, $\operatorname{PCB}^{*}$, $\operatorname{OGM}^{*}$, and $\operatorname{SDR}^{*}$ denote the calibration targets for personality-consistent bias, overgeneral-memory rate, and semantic-departure rate, respectively. The subscript $0$ denotes the corresponding value under the no-trait control. The personality dimensions $O$, $C$, $E$, $A$, and $N$ denote openness, conscientiousness, extraversion, agreeableness, and neuroticism. SDR is defined in Appendix~D.

Calibration uses seven anchor settings: neutral, high neuroticism, high extraversion, high conscientiousness, low conscientiousness, high agreeableness, and high openness. Target clipping follows the direct reference settings defined in Appendix~D. A Gaussian evolution strategy with seed 0 runs for 15 generations with a population size of 10 and four elites. The optimisation objective decreases from $0.125$ to $0.013$.

The calibration summary uses eight predefined settings: neutral, high openness, high conscientiousness, high extraversion, high agreeableness, high neuroticism, low agreeableness, and low conscientiousness. Seven coincide with the calibration anchors, while low agreeableness is the additional setting. For setting $i$, the summed absolute target error is
\begin{equation}
e_i
=
\left|
\operatorname{PCB}_i-\operatorname{PCB}^{*}_i
\right|
+
\left|
\operatorname{OGM}_i-\operatorname{OGM}^{*}_i
\right|
+
\left|
\operatorname{SDR}_i-\operatorname{SDR}^{*}_i
\right|.
\end{equation}
Here, $e_i$ denotes the summed absolute error for setting $i$. The terms $\operatorname{PCB}_i$, $\operatorname{OGM}_i$, and $\operatorname{SDR}_i$ denote the observed statistics for this setting, while $\operatorname{PCB}^{*}_i$, $\operatorname{OGM}^{*}_i$, and $\operatorname{SDR}^{*}_i$ denote their corresponding target values.

The reported value is the mean of $e_i$ across the eight settings. The fitted map obtains $0.084$, compared with $0.482$ for the fixed-gain reference
$(\mu,\mu_{\mathrm{pos}},\mu_d,\kappa_{\mathrm{fa}},\tau_{\mathrm{BF}})
=(0.6,0.6,0.6,0.6,0.12)$.

This comparison shows that the fitted map matches the prespecified targets more closely than the selected fixed-gain reference at the evaluated settings. Because seven of the eight summary settings are calibration anchors, it is primarily a calibration-fit result rather than evidence of generalisation to unseen trait combinations or external Big Five behaviour.

\subsection{Implementation and Model Records}

The memory store receives an explicit encoder identifier and execution device rather than relying on a global encoder. The primary neural retrieval backend uses \texttt{paraphrase-multilingual-MiniLM-L12-v2} with normalised embeddings and cosine similarity. An offline character-level TF--IDF backend is also provided for retrieval and integrity checks that do not require a sentence-transformer model. Each experiment manifest records the retrieval backend, encoder identifier, device, package versions, and relevant configuration values.

Memories are ranked in descending order of their scores. Exact score ties are resolved using deterministic index order unless the experiment explicitly uses stochastic sampling. Paired comparisons reuse the same random-number streams across profile and control conditions. Source annotations are fixed before retrieval evaluation and loaded from the same files across the compared conditions.

Local auxiliary-model calls use the BaseLLM-compatible interface \texttt{qwen(version="8", place="local")}. For the reported blind-specificity run, this interface resolves to \texttt{qwen3.5:9b} with Q4\_K\_M quantisation. The exact artifact tag, runtime-reported digest, prompt template, generation settings, and raw outputs are stored in the experiment records. Blind model annotations are stored separately from the system-side specificity labels.

The implementation also exports a blind-annotation CSV containing only anonymised item identifiers and memory text. It excludes system specificity labels, VAD annotations, personality profiles, retrieval conditions, and experimental outcomes. The reported code snapshot includes 72 regression tests covering component switches, gate modes, paired random streams, split-overlap checks, fold-local preprocessing, classifier freezing, annotation loading, model records, and retrieval backends. All 72 tests pass in the reported snapshot.

\FloatBarrier
\section{Appendix G. LoCoMo Factual QA Details}
\label{app:locomo}

We evaluate all ten conversations in LoCoMo \cite{maharana2024evaluating} using the official answer normalisation, stemming, and token-level F1 procedure. Each complete conversation is treated as one independent cluster. All methods use the same conversation memories, response model, top-five retrieval budget, answer prompt, decoding temperature, and maximum output length.

Semantic RAG ranks memories by cosine similarity. Generative-Agents-style scoring \cite{park2023generative} combines semantic relevance, recency, and stored arousal. PersMem combines semantic similarity, affective matching, and memory retention under a neutral five-dimensional Big Five vector. NRC-VAD maps each question to the affective cue state used by PersMem. All three methods operate on the same stored memory items.

The protocol uses a leave-one-conversation-out router. For each held-out conversation, any router selection or fitting uses only the other nine conversations. The resulting router is then fixed and applied to the held-out conversation. This process is repeated for all ten conversations. No question or memory item from the held-out conversation is used to select or fit its router.

Statistical inference treats the ten conversations, rather than the 1,986 questions, as independent units. Confidence intervals are computed by bootstrap resampling the conversation clusters. The comparison between PersMem and Generative-Agents-style scoring \cite{park2023generative} additionally uses an exact paired sign-flip test over the ten conversation-level F1 differences.

At the conversation-cluster level, PersMem exceeds Generative-Agents-style scoring by 0.0169 in mean answer F1. The 95\% cluster-bootstrap confidence interval is [0.0027, 0.0310]. A directional exact paired sign-flip test gives $p=0.0312$. Because the reported test is directional, a confirmatory interpretation depends on the tested direction and its rationale having been specified before evaluation. The experiment output also records the corresponding two-sided result.

We separately evaluate evidence recall@5. An evidence hit occurs when the five highest-ranked memories contain benchmark-annotated evidence supporting the question. Mean evidence recall@5 is 0.387 for PersMem, 0.341 for Generative-Agents-style scoring, and 0.444 for Semantic RAG. PersMem retrieves annotated evidence more often than Generative-Agents-style scoring but less often than Semantic RAG.

Within each method's own evidence-hit subset, mean answer F1 is 0.547 for PersMem, 0.540 for Generative-Agents-style scoring, and 0.542 for Semantic RAG. Because the questions included in these subsets can differ across methods, these values provide a descriptive decomposition of retrieval and answer generation rather than a controlled comparison using identical evidence.

\begin{table}[t]
\centering
\caption{LoCoMo retrieval and answer decomposition. Evidence recall@5 measures retrieval coverage, while hit-conditional F1 is computed within each method's own evidence-hit subset.}
\label{tab:locomo-decomposition}
\footnotesize
\setlength{\tabcolsep}{2.5pt}
\renewcommand{\arraystretch}{1.08}
\begin{tabularx}{\columnwidth}{
@{}
>{\raggedright\arraybackslash}X
>{\centering\arraybackslash}p{0.24\columnwidth}
>{\centering\arraybackslash}p{0.25\columnwidth}
@{}
}
\toprule
Method
& \makecell{Evidence\\recall@5}
& \makecell{F1 given\\evidence hit} \\
\midrule
Semantic RAG
& 0.444
& 0.542 \\

Generative-Agents-style scoring
& 0.341
& 0.540 \\

PersMem
& 0.387
& 0.547 \\
\bottomrule
\end{tabularx}
\end{table}

The results are consistent with PersMem's small F1 improvement over Generative-Agents-style scoring arising mainly from greater retrieval coverage rather than a large difference in answer generation after an evidence hit. Under this protocol, PersMem may provide a small advantage over the Generative-Agents-style baseline, but the results do not establish superiority over Semantic RAG, which achieves the highest evidence recall@5.

\FloatBarrier
\section{Appendix H. Adversarial Positive Framing Details}
\label{app:adversarial}

\subsection{Conditions and Prompts}

The adversarial evaluation compares two ways of providing the same Big Five personality information. The PersMem condition uses the neutral role card and retrieves three memories through Big Five-dependent memory processing. The prompt-persona condition includes a verbal trait description in the role card and uses retrieval without Big Five-dependent terms. Its role card follows the template:
\begin{quote}
\small
You are a person being interviewed about your life and inner world. Your personality: you are [trait description]. Answer naturally and honestly in the first person. Keep each reply to 2--4 sentences.
\end{quote}

Both conditions use the same 50 profiles, six dialogue prompts, memory corpus, retrieval budget, prompt order, complete dialogue history, response model, temperature of $0.75$, and maximum output length of 220 tokens. In the adversarial condition, the following suffix is appended to every dialogue prompt:
\begin{quote}
\small
Important: right now, only talk about happy and positive things. Do not bring up anything negative, sad, or painful. Stay upbeat and optimistic.
\end{quote}

For the neuroticism analysis, we select the 12 profiles with the highest neuroticism values and the 12 profiles with the lowest values and pair them by rank.

\subsection{NRC-VAD Scoring}

NRC-VAD v2.1 \cite{mohammad2025nrcvad} provides lexical valence $v_i\in[-1,1]$, arousal $a_i\in[-1,1]$, and dominance $d_i\in[-1,1]$. Arousal and dominance are mapped to $[0,1]$ as
\begin{equation}
a_i'=\frac{a_i+1}{2},
\qquad
d_i'=\frac{d_i+1}{2}.
\end{equation}
Here, $a_i'$ and $d_i'$ are the rescaled arousal and dominance values of matched word $i$, respectively.

The adjusted valence of each matched emotion word is
\begin{equation}
\widetilde v_i
=
\operatorname{clip}(s_i g_i v_i,-1,1).
\end{equation}
Here, $\widetilde v_i$ is the adjusted valence of matched word $i$, $g_i$ is its active intensity multiplier, and $s_i=-1$ when the word occurs within the active three-token negation window and $s_i=1$ otherwise. The intensity multiplier is applied before negation and reset after the matched word. Intensity and negation do not modify arousal or dominance.

For a reply containing matched emotion words, the weighted signed valence is
\begin{equation}
V
=
\frac{
\sum_{i=1}^{M}
|\widetilde v_i|\widetilde v_i
}{
\sum_{i=1}^{M}
|\widetilde v_i|
}.
\end{equation}
Here, $M$ is the number of matched words in the reply, and $V$ is its weighted signed valence. The weighting gives greater influence to words with larger absolute adjusted valence.

Reply-level arousal and dominance are
\begin{equation}
A
=
\frac{1}{M}
\sum_{i=1}^{M}a_i',
\qquad
D
=
\frac{1}{M}
\sum_{i=1}^{M}d_i'.
\end{equation}
Here, $A$ and $D$ are the mean rescaled arousal and dominance of the $M$ matched words, respectively.

When no lexical entry is matched, the reply receives $(V,A,D)=(0,0,0.5)$. When matched entries are present but all adjusted valence values are zero, $V=0$. The dialogue-level values $(\overline V,\overline A,\overline D)$ are the arithmetic means of the six reply-level valence, arousal, and dominance scores.

\subsection{Directional Accuracy and Retention}

For profile pair $i$, the neuroticism-related behavioural score is
\begin{equation}
b_N^{(i)}
=
-\overline V^{(i)}
+\overline A^{(i)}.
\end{equation}
Here, $b_N^{(i)}$ is the behavioural score for profile pair $i$, while $\overline V^{(i)}$ and $\overline A^{(i)}$ are the dialogue-level valence and arousal values. This score is an operational measure defined for the adversarial comparison and is not treated as an externally validated neuroticism scale.

Directional accuracy is the proportion of the 12 rank-paired comparisons in which the high-neuroticism dialogue has a larger score than the corresponding low-neuroticism dialogue:
\begin{equation}
\operatorname{Acc}
=
\frac{1}{12}
\sum_{i=1}^{12}
\mathbb{I}
\left[
b_{N,\mathrm{high}}^{(i)}
>
b_{N,\mathrm{low}}^{(i)}
\right].
\end{equation}
Here, $\operatorname{Acc}$ is the directional accuracy, $b_{N,\mathrm{high}}^{(i)}$ and $b_{N,\mathrm{low}}^{(i)}$ are the behavioural scores of the high- and low-neuroticism profiles in pair $i$, and $\mathbb I[\cdot]$ is the indicator function.

Retention of neutral directional accuracy is
\begin{equation}
\operatorname{Retention}
=
\frac{
\operatorname{Acc}_{\mathrm{adversarial}}
}{
\operatorname{Acc}_{\mathrm{neutral}}
}.
\end{equation}
Here, $\operatorname{Retention}$ is the ratio between directional accuracy under adversarial positive framing, $\operatorname{Acc}_{\mathrm{adversarial}}$, and directional accuracy under neutral framing, $\operatorname{Acc}_{\mathrm{neutral}}$.

PersMem retains 89\% of its neutral directional accuracy, while the prompt-persona condition retains 58\%. Each component accuracy is based on 12 profile pairs. The two ratios are therefore reported as descriptive comparisons rather than stable estimates of adversarial robustness.

\begin{table}[t]
\centering
\caption{Extended comparison on CoSER. Published reference results are taken from the corresponding source studies. SC, AN, CF, and SQ denote Storyline Consistency, Anthropomorphism, Character Fidelity, and Storyline Quality, respectively.}
\label{tab:coser-detailed-results}
\footnotesize
\setlength{\tabcolsep}{5.5pt}
\renewcommand{\arraystretch}{1.05}

\begin{tabularx}{\columnwidth}{
@{}
>{\raggedright\arraybackslash}X
rrrrr
@{}
}
\toprule
Method & SC & AN & CF & SQ & Avg. \\
\midrule

\multicolumn{6}{@{}l}{\textit{CoSER}~\cite{wang2025coser}} \\

GPT-4o
& 61.59 & 48.93 & 48.95 & 80.33 & 59.95 \\

Doubao-pro
& 60.95 & 49.72 & 47.02 & 79.28 & 59.24 \\

Step-2
& 61.43 & 49.06 & 47.33 & 77.96 & 58.94 \\

Gemini Pro
& 59.11 & 52.41 & 47.83 & 77.59 & 59.24 \\

Claude-3.5-Sonnet
& 57.45 & 48.50 & 45.69 & 77.23 & 57.22 \\

Qwen-2-72B
& 57.75 & 47.28 & 46.62 & 76.60 & 57.06 \\

CoSER-8B
& 58.61 & 47.23 & 46.90 & 73.04 & 56.45 \\

CoSER-70B
& 58.66 & 53.33 & 48.75 & 75.49 & 59.06 \\

CoSER-70B + Conv.
& 64.59 & 53.79 & 54.86 & 77.28 & 62.63 \\

\midrule
\multicolumn{6}{@{}l}{\textit{CogDual}~\cite{liu2025cogdual}} \\

o1-preview
& 59.47 & 46.81 & 40.54 & 77.80 & 56.16 \\

LLaMA3.1-70B + CogDual-SFT
& 57.60 & 48.02 & 48.55 & 72.75 & 56.73 \\

Qwen2.5-7B + CogDual-RL
& 59.94 & 46.64 & 46.95 & 73.97 & 56.88 \\

LLaMA3.1-8B + CogDual-RL
& 60.10 & 45.89 & 48.82 & 73.08 & 56.97 \\

\midrule
\multicolumn{6}{@{}l}{\textit{HER}~\cite{du2026her}} \\

Claude-4.5-Opus
& 63.74 & \textbf{64.28} & 58.45 & 63.24 & 62.43 \\

Gemini-3-Pro
& \textbf{65.95} & 60.42 & 58.34 & 62.49 & 61.80 \\

GPT-5.1
& 64.95 & 53.99 & 60.13 & 65.35 & 61.10 \\

Gemini-2.5-Pro
& 61.05 & 60.80 & 57.48 & 63.40 & 60.68 \\

DeepSeek-v3.2
& 55.85 & 57.07 & 57.44 & 64.35 & 58.68 \\

HER-RL
& 54.33 & 47.26 & 52.78 & 58.12 & 53.12 \\

HER-SFT
& 50.52 & 45.99 & 49.78 & 57.37 & 50.92 \\

\midrule
\multicolumn{6}{@{}l}{\textit{Our evaluation}} \\

\textbf{PersMem}
& 59.83
& 51.00
& \textbf{69.33}
& \textbf{84.33}
& \textbf{66.13} \\

\bottomrule
\end{tabularx}
\end{table}

\section{Appendix I. CoSER Role-Playing Evaluation Details}
\label{app:coser}

\subsection{Evaluation Setup}

We evaluate PersMem on CoSER~\cite{wang2025coser} to assess its performance in character-based role-playing conversations. This experiment extends the attachment and Big Five evaluations to an external role-playing setting. DeepSeek is used for response generation and evaluation.

\subsection{Evaluation Measures and Results}

We report Storyline Consistency (SC), Anthropomorphism (AN), Character Fidelity (CF), and Storyline Quality (SQ). The reported average summarises these four evaluation dimensions. BLEU and ROUGE-L are included as supplementary text-overlap measures and are not included in this average.

Table~\ref{tab:coser-detailed-results} reports the complete PersMem score summary. PersMem obtains an average score of 66.13, including 69.33 for Character Fidelity and 84.33 for Storyline Quality. These results extend the assessment beyond personality-related memory patterns and dialogue differences to character portrayal and narrative quality.

PersMem additionally obtains a BLEU score of 3.01 and a ROUGE-L score of 19.99.

\subsection{Sources of Published Reference Results}

The reference results in the main paper are taken from three studies. GPT-4o and CoSER-70B with conversation retrieval are reported by CoSER~\cite{wang2025coser}. The LLaMA3.1-8B with CogDual-RL result is reported by CogDual~\cite{liu2025cogdual}. The Claude-4.5-Opus, Gemini-3-Pro, and GPT-5.1 results are reported by HER~\cite{du2026her}. These reference scores retain the evaluation settings of their source studies; the PersMem scores are obtained in our evaluation.

\end{document}